\pdfoutput=1

\documentclass[11pt]{article}

\usepackage[final]{acl}

\usepackage{times}
\usepackage{latexsym}
\usepackage{booktabs} 
\usepackage{fontawesome}
\usepackage{colortbl}
\usepackage{xcolor}
\usepackage[T1]{fontenc}

\usepackage[utf8]{inputenc}

\usepackage{microtype}

\usepackage{inconsolata}

\usepackage{graphicx}

\usepackage{bibentry}
\usepackage{color}
\usepackage{multirow}
\usepackage{colortbl}
\usepackage{tabularx}
\usepackage{textgreek}
\usepackage{csquotes}
\usepackage{amsmath}
\usepackage{amssymb}
\usepackage{enumitem}
\usepackage{hyperref}
\definecolor{mygray}{gray}{0.95}
\title{Pandora's Box or Aladdin's Lamp: A Comprehensive Analysis Revealing the Role of RAG Noise in Large Language Models}

\author{
  \textbf{Jinyang Wu}\textsuperscript{\rm 1}, \textbf{Shuai Zhang}\textsuperscript{\rm 1}\thanks{\quad Corresponding Authors}, \textbf{Feihu Che}\textsuperscript{\rm 2}, \textbf{Mingkuan Feng}\textsuperscript{\rm 1}\\
  \textbf{Chuyuan Zhang}\textsuperscript{\rm 1}\quad
  \textbf{Pengpeng Shao}\textsuperscript{\rm 1}, \textbf{Jianhua Tao}\textsuperscript{\rm 1,2}\footnotemark[1]\\
  \textsuperscript{\rm 1}Department of Automation, Tsinghua
University\\
  \textsuperscript{\rm 2}Beijing National Research Center for
Information Science and Technology\\
    \texttt{wu-jy23@mails.tsinghua.edu.cn, zhang$\_$shuai@mail.tsinghua.edu.cn}\\
    }

\begin{document}
\maketitle
\begin{abstract}
Retrieval-Augmented Generation (RAG) has emerged as a key method to address hallucinations in large language models (LLMs). While recent research has extended RAG models to complex noisy scenarios, these explorations often confine themselves to limited noise types and presuppose that noise is inherently detrimental to LLMs, potentially deviating from real-world retrieval environments and restricting practical applicability. In this paper, we define seven distinct noise types from a linguistic perspective and establish a Noise RAG Benchmark (NoiserBench), a comprehensive evaluation framework encompassing multiple datasets and reasoning tasks. Through empirical evaluation of eight representative LLMs with diverse architectures and scales, we reveal that these noises can be further categorized into two practical groups: noise that is beneficial to LLMs (aka beneficial noise) and noise that is harmful to LLMs (aka harmful noise). While harmful noise generally impairs performance, beneficial noise may enhance several aspects of model capabilities and overall performance. Our analysis offers insights for developing robust RAG solutions and mitigating hallucinations across diverse retrieval scenarios. Code is available at \href{https://github.com/jinyangwu/NoiserBench}{https://github.com/jinyangwu/NoiserBench}.
\end{abstract}

\section{Introduction}\label{sec1}
\begin{figure}[ht]
\centering
\includegraphics[width=0.95\columnwidth]{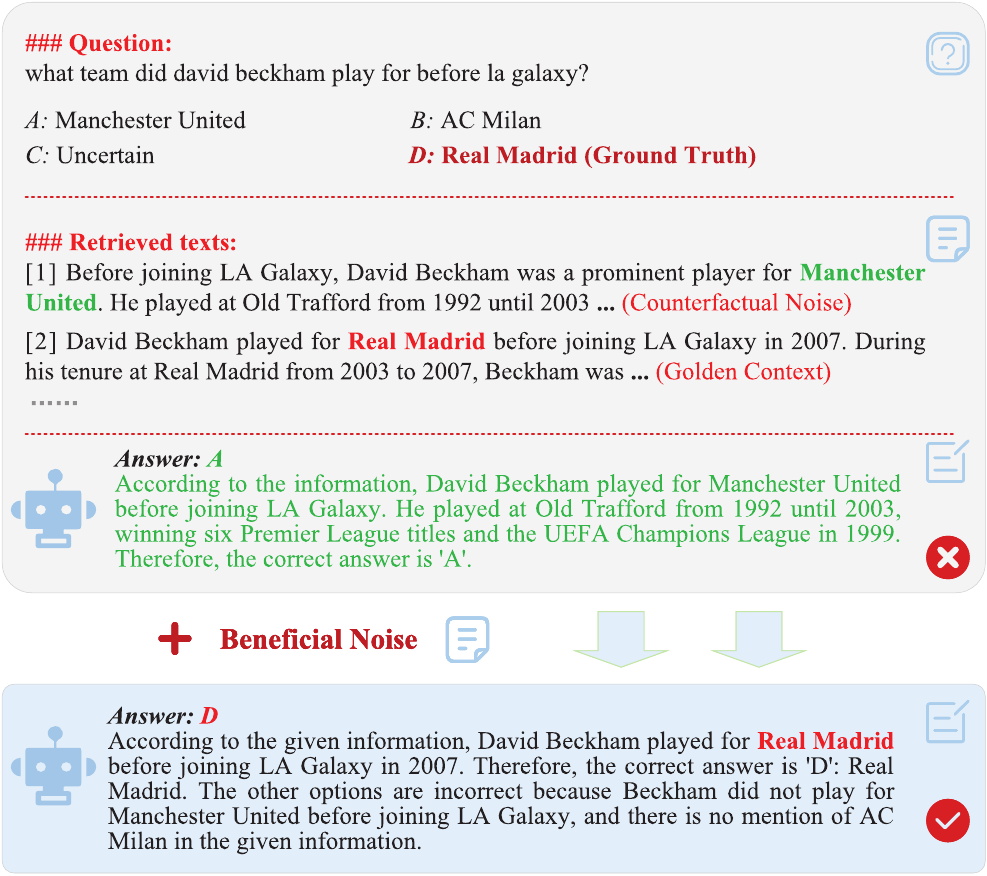} 
\caption{An example from NoiserBench illustrating effects of different RAG noises. Initially, the model is misled by counterfactual noise. Interestingly, upon introducing beneficial noise, it successfully discriminates between correct and incorrect information and produces the accurate answer `D'.}
\vskip -0.15in
\label{Figure1}
\end{figure}

Large language models (LLMs)~\citep{open2023introducing,meta2024introducing} have demonstrated remarkable proficiency across various tasks~\citep{bubeck2023sparks}. Despite impressive capabilities, LLMs face challenges such as reliance on outdated knowledge and hallucination~\citep{huang2023survey,kandpal2023large}. Retrieval-Augmented Generation (RAG) has recently emerged as a promising approach to mitigate these limitations~\citep{lewis2020retrieval,gao2024retrievalaugmentedgenerationlargelanguage}. RAG enhances LLMs' performance by augmenting inputs with additional information retrieved from external sources during inference.

However, external sources often contain various non-standard noises, including fake news, outdated content, spelling errors, and data contamination, which may potentially influence model performance~\citep{shi2023large,xie2024adaptivechameleonstubbornsloth}. It is crucial to explore how noise affects RAG systems and understand the underlying mechanisms.

Recent studies~\citep{chen2024benchmarking,xiang2024certifiablyrobustragretrieval} have attempted to extend RAG systems to complex real-world scenarios, investigating the impact of noisy documents and strategies to enhance the system's robustness. For example, \citet{cuconasu2024power} defines three types of noise in retrieved documents and examines their impacts on LLMs. Despite highlighting one noise's positive effect, the study lacks a comprehensive noise definition and in-depth investigation of underlying principles. \citet{fang2024enhancingnoiserobustnessretrievalaugmented} applies adversarial training to dynamically adjust the model’s training process in response to retrieval noises. RobustRAG~\citep{xiang2024certifiablyrobustragretrieval} proposes a defense framework against retrieval corruption attacks. Nevertheless, these investigations typically focus on a limited number of noise types (usually no more than three) and lack clear classification, which fails to fully capture the complexity of real-world noise environments. Additionally, these studies often assume that noise is harmful, neglecting its potential positive effects and lacking systematic evaluation datasets. As shown in Figure \ref{Figure1}, introducing beneficial noise allows the LLMs to avoid the harmful effects of counterfactual noise, focus on the golden context, and produce accurate responses. Thus, this highlights the urgent need for systematic noise taxonomy and comprehensive evaluation of retrieval noise impacts in RAG systems.

In this paper, we comprehensively analyze the role of RAG noise in LLMs. We first define seven types of noise from a linguistic perspective. Based on this definition, we propose a systematic framework to create diverse noisy documents and establish NoiserBench, a novel noise RAG benchmark. Then, we evaluate eight representative LLMs with different architectures and scales. Extensive results show that RAG noises can be categorized into two practical groups: \textit{beneficial noise} (semantic, datatype, illegal sentence) and \textit{harmful noise} (counterfactual, supportive, orthographic, prior). While harmful noise impairs performance, beneficial noise surprisingly enhances model capabilities and leads to improved performance. Further analysis reveals that beneficial noise facilitates more standardized answer formats, clearer reasoning paths, and increases confidence in responses with golden context. These contrasting effects are analogous to \textit{opening Pandora's Box} (harmful noise) versus \textit{unlocking Aladdin's Lamp} (beneficial noise). This study aims to advance research on mitigating harmful noise while leveraging beneficial noise effects. Our main contributions are:
\begin{itemize}
    \item We define seven types of noise and categorize them into two groups: beneficial and harmful noise. This is the first comprehensive study to define and assess RAG noises from both linguistic and practical perspectives.
    \item We introduce a novel framework for constructing diverse retrieval documents and create NoiserBench, a benchmark that effectively simulates real-world noise in RAG models.
    \item Evaluated on multiple datasets and LLMs, our results reveal that while some RAG noises (e.g. counterfactual) can open Pandora's Box and cause errors, beneficial noise (e.g. datatype) has the potential to unlock the power of Aladdin's Lamp and deliver positive effects. 
    \item Our findings redefine retrieval noise and encourage researchers to explore methods that harness its beneficial properties while addressing its harmful effects.
\end{itemize}

\section{Related Work}\label{sec2}
\textbf{Retrieval-Augmented Generation}\;
By integrating external information, RAG methods enhance reasoning and generation process~\citep{gao2024retrievalaugmentedgenerationlargelanguage,zhao2024retrievalaugmentedgenerationaigeneratedcontent}. Early works primarily focus on improving retrieval model performance to obtain relevant documents for subsequent generation~\citep{qu-etal-2021-rocketqa,wang2023selfknowledgeguidedretrievalaugmentation,zheng2024ksllmknowledgeselectionlarge}. Recent research has expanded RAG framework to real-world noisy scenarios, aiming to build robust RAG systems by enhancing the generator~\citep{fang2024enhancingnoiserobustnessretrievalaugmented,xiang2024certifiablyrobustragretrieval}. For instance, Self-RAG~\citep{asai2024selfrag} employs four specialized tokens and GPT-4-generated instruction-tuning data to fine-tune the Llama2 model. RobustRAG~\citep{xiang2024certifiablyrobustragretrieval} proposes an isolate-then-aggregate defense framework to enhance model robustness against retrieval corruption attacks. However, these investigations are constrained by their narrow focus on specific noise types and the inherent assumption that noise is harmful, potentially hindering method generalization. This paper aims to analyze RAG noise and reveal its roles systematically.\\

\begin{figure*}[ht]
\centering
\includegraphics[width=1.00\linewidth]{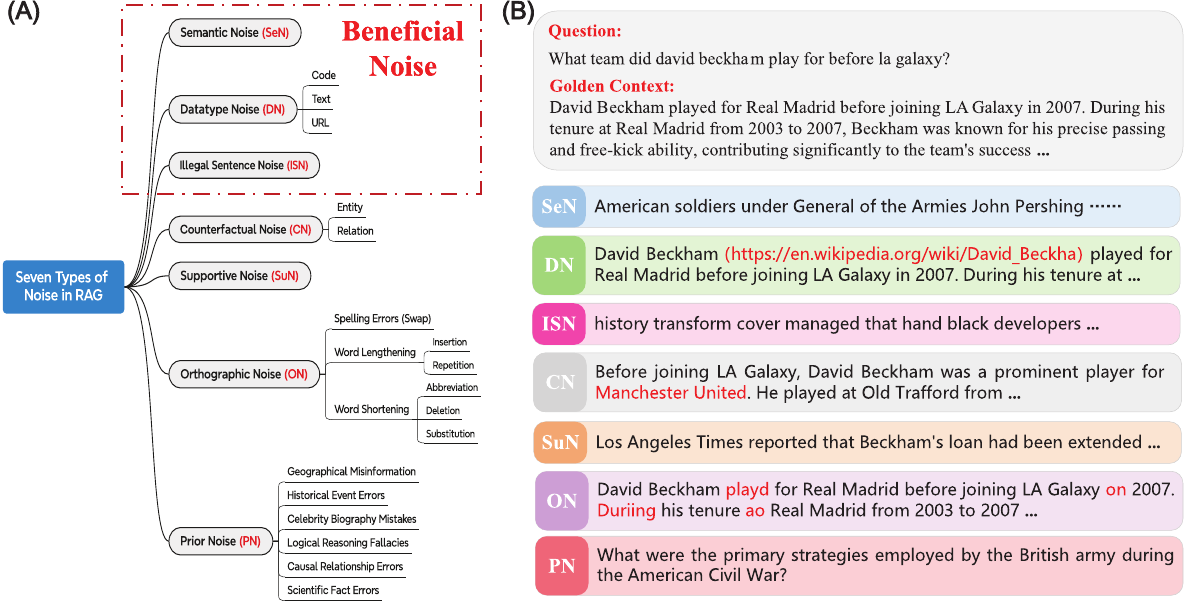}
\caption{\textbf{(A)} Our seven RAG noise types comprehensively capture real-world retrieval challenges. \textbf{(B)} This detailed illustration intuitively depicts the diverse RAG noise landscape, with noise injection regions marked in \textcolor{red}{red}.}
\label{Figure2}
\end{figure*}

\vskip -0.1in
\noindent\textbf{Noise Injection in LLMs}\;
Noise injection~\citep{grandvalet1997noise} in LLMs involves adding noise to inputs during training or inference, such as data augmentation~\citep{ye2024llmdadataaugmentationlarge}, adversarial training~\citep{fang2024enhancingnoiserobustnessretrievalaugmented}, and prompt perturbation~\citep{zhu2024promptrobustevaluatingrobustnesslarge}. Recently, researchers have focused on noise injection in RAG systems~\citep{chen2024benchmarking}. For example, \citet{cuconasu2024power} classifies three retrieval noises and explores their effects on LLMs. \citet{fang2024enhancingnoiserobustnessretrievalaugmented} leverages adversarial training to dynamically adjust LLMs' training process in response to retrieval noises. However, these noise types are limited to reflect complex real-world scenarios. A comprehensive framework that simulates real-world noise is necessary.

\section{A Taxonomy of RAG Noise}\label{sec3}
As shown in Figure \ref{Figure2}, we categorize RAG noise into seven linguistic types. They are further divided into beneficial (semantic, datatype, and illegal sentence) and harmful noise (counterfactual, supportive, orthographic, and prior) for practical applications. We will explain the reason behind this classification in \textit{\ref{sec5} Experiment Setup}.

\vspace{0.3em}
\textbf{Semantic Noise (SeN)}\;
Retrieval documents may contain content with low semantic relevance to the query, often being off-topic or deviating from the intended meaning. Given that Warren Weaver originally defined semantic noise as "the perturbations or distortions of sentence meaning"~\citep{shannon1961mathematical}, we classify off-topic, low-semantic-relevance documents as \textit{semantic noise}.

\vspace{0.3em}
\textbf{Datatype Noise (DN)}\;
This type of noise refers to the mixing of different data types on the web, such as the blending of links and text on Wikipedia. In this paper, we consider three data types: text, URLs, and code.

\vspace{0.3em}
\textbf{Illegal Sentence Noise (ISN)}\;
Web content may include fragments that do not form grammatically correct sentences, such as ``history transform cover managed that hand black''. We define this type of noise as \textit{illegal sentence noise}.

\vspace{0.3em}
\textbf{Counterfactual Noise (CN)}\;
The internet contains abundant false information, including fake news and outdated knowledge~\citep{tumarkin2001news,olan2024fake}, presenting critical challenges to RAG systems. Drawing from linguistics, where ``counterfactual" denotes statements contrary to fact~\citep{feng2006if}, we introduce the term ``\textit{counterfactual noise}" to characterize factual errors. This concept aligns with prior research~\citep{fang2024enhancingnoiserobustnessretrievalaugmented}.

\vspace{0.3em}
\textbf{Supportive Noise (SuN)}\;
Supportive evidence, known as positive evidence, is highly semantically relevant to a hypothesis and provides necessary information to support it~\citep{kertesz2012data}. We introduce the term ``\textit{supportive noise}'' to describe documents that exhibit high semantic relevance but lack corresponding answer information.

\vspace{0.3em}
\textbf{Orthographic Noise (ON)}\;
The word ``orthography'' originates from the Greek \textit{orthós} (meaning ``correct'') and \textit{gráphein} (meaning ``to write''), and refers to the way words are written in linguistics~\citep{skeat1993concise,aloufi2021language}. \textit{Orthographic noise}, on the other hand, can refer to writing errors such as spelling mistakes and word lengthening.

\begin{figure*}[ht]
\centering
\includegraphics[width=1.00\linewidth]{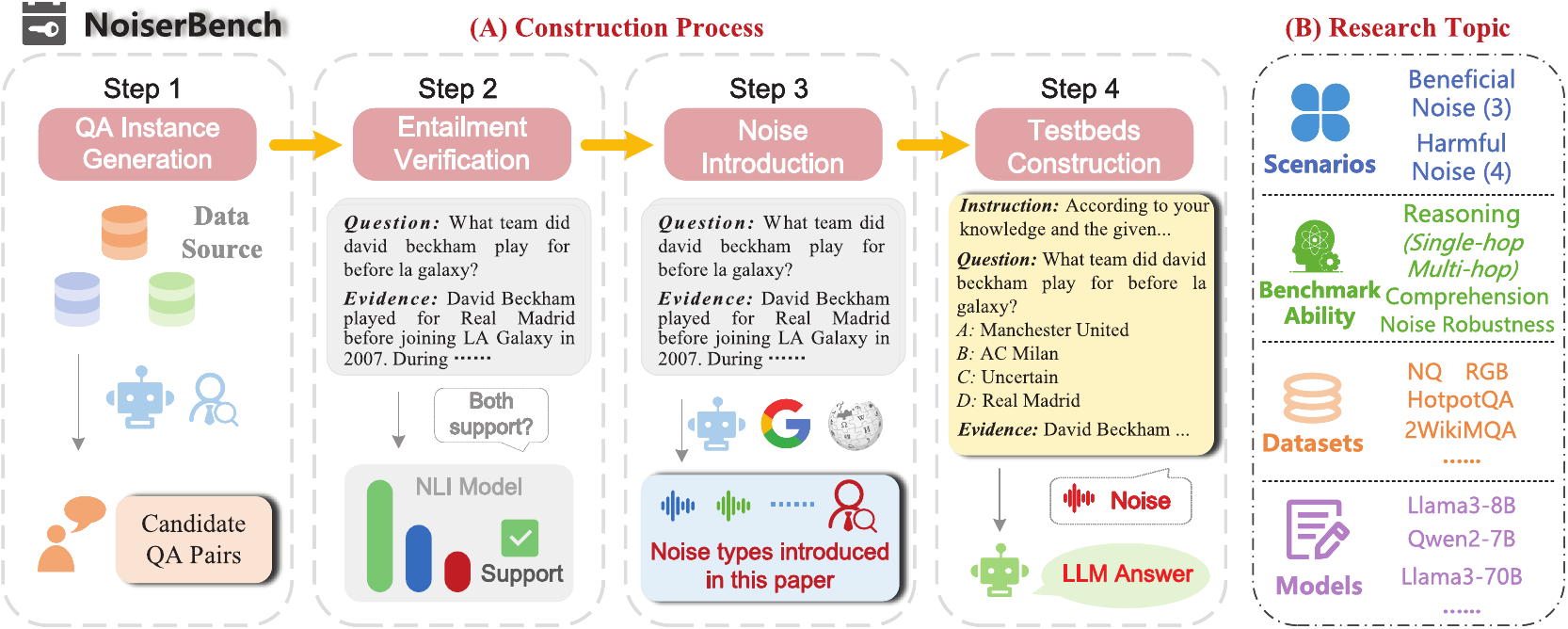}
\caption{The overall framework for simulating the impact of real-world noise on RAG models. Initially, we generate and obtain QA instances, utilizing ChatGPT to filter out ambiguous examples (\textbf{\textit{Step 1}}). Then, we perform entailment verification using NLI models to maintain evidence quality (\textbf{\textit{Step 2}}). After that, we use tools like search engines to create noisy documents (\textbf{\textit{Step 3}}). Finally, we transform the free-form QA into a multiple-choice QA format by providing several answer options for convenient automatic evaluation (\textbf{\textit{Step 4}}). All experiments are conducted in a zero-shot setting to avoid bias from demonstrations.}
\label{Figure3}
\end{figure*}

\vspace{0.3em}
\textbf{Prior Noise (PN)}\;
In linguistics, prior knowledge refers to what a learner already knows before solving a problem~\citep{chafe1971linguistics}. Our study defines \textit{prior noise} as questions based on false assumptions or premises. For example, the question ``Who was the CEO of Google when they were restructured into Alphabet in 2017?'' contains prior noise because the restructuring occurred in 2015, not 2017.

\section{Noise RAG Benchmark Construction}\label{sec4}
The overall framework is illustrated in Figure \ref{Figure3}. We will discuss the data construction and evaluation metrics as follows.

\subsection{Data Construction}
As shown in Figure \ref{Figure3} (A), our framework comprises four essential steps, including QA Instance Generation, Entailment Verification, Noise Introduction and Testbeds Construction.

\vspace{0.5em}\noindent
\textbf{Step 1: QA Instance Generation}\;
For prior noise, we collect article snippets from mainstream media and Wikipedia, covering various time periods and domains such as sports, politics, and finance. We then design prompts for ChatGPT to generate relevant events, questions, and answers for each snippet. Note that the generated questions contain prior noise (factual errors), which we manually review to ensure that they are reasonably answerable by LLMs. For the remaining seven types of noise (SeN, DN, ISN, CN, SuN, ON, PN), we obtain question-answering (QA) pairs from existing datasets, following previous work~\citep{fang2024enhancingnoiserobustnessretrievalaugmented,cuconasu2024power,yoran2024making}. After obtaining candidate QA pairs, we employ ChatGPT to remove ambiguous or difficult-to-assess pairs, followed by a manual review. For example, questions like ``How many companies have a market capitalization of over $\$$25 billion and pledged to reduce greenhouse gas emissions?'' should be excluded due to their broad potential answers and the dynamic market values of companies. Similar criteria are applied to other instances.

\vspace{0.5em}\noindent
\textbf{Step 2: Entailment Verification}\;
As illustrated in \citet{xie2024adaptivechameleonstubbornsloth,yoran2024making}, effective evidence should strongly support its answer. For example, golden evidence about David Beckham should support that he played for Real Madrid before joining LA Galaxy. Therefore, we employ the natural language inference model, bart-large-mnli-407M~\citep{lewis2019bartdenoisingsequencetosequencepretraining} to ensure evidence properly entails the answer. We only keep those examples with an entailment probability $ p \ge 0.8$.

\vspace{0.5em}\noindent
\textbf{Step 3: Noise Introduction}\;
We construct diverse retrieval documents for noise testbeds. For counterfactual noise, we extract related entities and relations from Google search results to create counterfactual answers. ChatGPT is then employed to construct corresponding supportive evidence, followed by entailment verification. For Supportive and semantic noise, we utilize the 2018 English Wikipedia dump~\citep{karpukhin2020densepassageretrievalopendomain} as source documents, with off-the-shelf Contriever-MS MARCO model~\citep{izacard2022unsuperviseddenseinformationretrieval} for retrieval and the lightweight text embedding model all-MiniLM-L6-v2~\citep{wang2021minilmv2multiheadselfattentionrelation} for semantic relevance filtering. To simulate illegal sentence noise, we construct meaningless sentences by randomly combining words from model vocabulary, mimicking real-world garbled text. Datatype noise is created by prompting ChatGPT to insert URLs or code snippets while preserving key answer information. Finally, orthographic noise is generated using the open-source textnoisr package~\citep{textnoisr}. This pipeline enables a comprehensive assessment of model performance across a range of noise scenarios.

\vspace{0.6em}\noindent
\textbf{Step 4: Testbeds Construction}\;
After obtaining high-quality QA instances and diverse retrieval documents, we build testbeds to evaluate model performance under various noise conditions. Given the challenges in automatically assessing LLMs' responses to open-ended QA tasks~\citep{xie2024adaptivechameleonstubbornsloth}, we convert free-form QA into a multiple-choice format. This constrains the response space and facilitates more accurate evaluation. Specifically, for each QA pair, LLMs choose from 4 options: the correct answer, two counterfactual alternatives, and ``Uncertain''. The order of the golden option remains entirely random to avoid LLMs' sensitivity to option order~\citep{wu2024largelanguagemodelsunderstand}. 

Finally, eight datasets are obtained for NoiserBench. Following prior works~\citep{yoran2024making,wang2024ratretrievalaugmentedthoughts}, we randomly select 500 samples from each dataset as test cases or use all samples if the size of this dataset is smaller than 500.

\subsection{Evaluation Metrics}
This benchmark aims to reveal the role that RAG noise plays on LLMs. We use accuracy as the primary metric and also report the weighted average accuracy across datasets.

\begin{table*}[ht]
\centering
\caption{Impact of diverse noise types on accuracy ($\%$) for Llama3-8B-Instruct and Qwen2-7b-Instruct across seven datasets. We assess performance across various retrieval scenarios: ``Base'' (no retrieval), ``Golden Only'' (only golden retrieval context), and ``Golden $\&$ XXX" (golden context + specific retrieval noises, including Counterfactual, Supportive, Orthographic, Semantic, Datatype, Illegal Sentence Noise). The \textcolor{green}{green} and \textcolor{red}{red} values indicate the performance gap from "Golden Only". We also provide the weighted average accuracy for each noise type. The best two results are shown in bold and underlined.}\label{table1}
\resizebox{1.0\linewidth}{!}{
\begin{tabular}{lllllllll} 
\toprule
\multicolumn{9}{c}{Llama3-8B-Instruct} \\
\midrule
\multirow{2.5}{*}{{\bfseries Scenario}} & \multicolumn{2}{c}{{\bfseries Single-hop}} & \multicolumn{3}{c}{{\bfseries Multi-hop (Explicit)}} & \multicolumn{2}{c}{{\bfseries Multi-hop (Implicit)}} & \multirow{2.5}{*}{{\bfseries Average}}\\
\cmidrule(r){2-3} \cmidrule(r){4-6} \cmidrule(r){7-8}
\quad & NQ & RGB & HotpotQA & 2WikiMQA & Bamboogle & StrategyQA & TempQA   \\
\midrule
Base & 61.34 & 47.00 & 53.80 & 34.40 & 32.00 & 58.80 & 50.54 & 51.58 \\
Golden Only & 93.06 & 80.00 & 97.80 & 79.80 & 87.20 & \underline{73.40} & 91.94 & 86.57 \\
Golden \& CN & 58.86 & 36.33 & 44.20 & 21.20 & 61.60 & 43.20 & 67.74 & $45.58_{\textcolor{green}{-40.99}}$ \\
Golden \& SuN & 90.58 & 80.00 & 95.60 & 81.00 & 93.60 & 69.40 & 93.01 & $85.37_{\textcolor{green}{-1.20}}$ \\
Golden \& ON & 93.31 & 75.00 & 96.20 & 78.60 & 89.60 & 63.60 & 90.86 & $83.99_{\textcolor{green}{-2.58}}$ \\
\rowcolor{mygray}
Golden \& SeN & \underline{96.53}$_{\textcolor{red}{+0.47}}$ & 81.33$_{\textcolor{red}{+1.33}}$ & \underline{98.40}$_{\textcolor{red}{+0.60}}$ & \underline{87.20}$_{\textcolor{red}{+7.40}}$ & \underline{93.60}$_{\textcolor{red}{+6.40}}$ & 68.40 & \underline{96.24}$_{\textcolor{red}{+4.30}}$ & $88.73_{\textcolor{red}{+2.16}}$ \\
\rowcolor{mygray}
Golden \& DN & 93.19$_{\textcolor{red}{+0.13}}$ & \underline{81.67}$_{\textcolor{red}{+1.67}}$ & 95.00 & 82.00$_{\textcolor{red}{+2.20}}$ & 88.00$_{\textcolor{red}{+0.80}}$ & \bf{73.60}$_{\textcolor{red}{+0.20}}$ & 94.62$_{\textcolor{red}{+2.68}}$ & $86.91_{\textcolor{red}{+0.34}}$ \\
\rowcolor{mygray}
Golden \& ISN & {\bfseries 96.65}$_{\textcolor{red}{+0.65}}$ & {\bfseries 83.00}$_{\textcolor{red}{+1.33}}$ & {\bfseries 98.80}$_{\textcolor{red}{+1.00}}$ & {\bfseries 87.40}$_{\textcolor{red}{+7.60}}$ & {\bfseries 94.40}$_{\textcolor{red}{+7.20}}$ & 72.60 & {\bfseries 97.85}$_{\textcolor{red}{+5.91}}$ & {\bfseries 89.89}$_{\textcolor{red}{+3.32}}$ \\

\midrule

\multicolumn{9}{c}{Qwen2-7B-Instruct} \\
\midrule
Base & 58.24 & 31.33 & 50.20 & 22.60 & 31.20 & 42.40 & 40.86 & 43.01 \\
Golden Only & {\bfseries 97.03} & 76.33 & 98.40 & 78.00 & 94.40 & \underline{67.00} & 94.62 & 86.46 \\
Golden \& CN & 41.88 & 26.00 & 38.40 & 12.40 & 39.20 & 37.60 & 45.16 & $33.96_{\textcolor{green}{-52.50}}$ \\
Golden \& SuN & 90.46 & 74.00 & 96.40 & \underline{80.40} & 92.00 & 64.00 & 90.32 & $83.65_{\textcolor{green}{-2.81}}$ \\
Golden \& ON & 95.66 & 74.00 & 97.80 & 80.00 & 91.20 & 54.60 & 94.62 & $83.82_{\textcolor{green}{-2.64}}$ \\
\rowcolor{mygray}
Golden \& SeN & 96.53 & 77.67$_{\textcolor{red}{+1.34}}$ & \underline{98.80}$_{\textcolor{red}{+0.40}}$ & 77.00 & {\bfseries 96.80}$_{\textcolor{red}{+2.40}}$ & 66.80 & \underline{97.31}$_{\textcolor{red}{+2.69}}$ & $86.60_{\textcolor{red}{+0.14}}$ \\
\rowcolor{mygray}
Golden \& DN & 96.03 & {\bfseries 84.33}$_{\textcolor{red}{+9.00}}$ & 98.20 & 79.60$_{\textcolor{red}{+1.60}}$ & 93.60 & {\bfseries 71.80}$_{\textcolor{red}{+4.80}}$ & 95.70$_{\textcolor{red}{+1.08}}$ & {\bfseries 88.11}$_{\textcolor{red}{+1.65}}$ \\
\rowcolor{mygray}
Golden \& ISN & \underline{96.65} & \underline{80.00}$_{\textcolor{red}{+3.67}}$ & {\bfseries 99.00}$_{\textcolor{red}{+0.60}}$ & {\bfseries 83.80}$_{\textcolor{red}{+5.80}}$ & {\bfseries 96.80}$_{\textcolor{red}{+2.40}}$ & 66.80 & {\bfseries 97.85}$_{\textcolor{red}{+1.23}}$ & {\bfseries 88.11}$_{\textcolor{red}{+1.65}}$ \\
\bottomrule
\end{tabular}}
\end{table*}

\section{Experiment Setup}\label{sec5}
\subsection{Datasets}\label{sec5.1}
We experiment with multiple QA datasets, which are categorized into four types based on the required reasoning skills:
\begin{itemize}
    \item {\bfseries Single-hop}: Questions requiring one-step reasoning. We evaluate using the Natural Questions (NQ)~\citep{kwiatkowski2019natural} and RGB~\citep{chen2024benchmarking} datasets.
    \item {\bfseries Explicit Multi-hop}: Questions where multiple reasoning steps are explicitly expressed. We utilize HotpotQA~\citep{yang2018hotpotqa}, 2WIKIMQA~\citep{welbl2018constructing} and Bamboogle dataset~\citep{press2022measuring}.
    \item {\bfseries Implicit Multi-hop}: Questions where intermediate steps are not explicitly stated, often requiring commonsense knowledge for implicit reasoning. We use StrategyQA~\citep{geva2021did} and TempQA~\citep{jia2018tempquestions}.
    \item {\bfseries Mixed-Hop}: Questions requiring single- or multi-hop reasoning. We use our constructed dataset, PriorQA.
\end{itemize}

\subsection{Baseline Models}\label{sec5.2}
We evaluate eight LLMs of different architectures and scales: Llama3-Instruct (8B, 70B)~\citep{meta2024introducing}, Qwen2-7B-Instruct~\citep{yang2024qwen2technicalreport}, Mistral (7B, 8x7B)~\citep{jiang2023mistral7b, jiang2024mixtralexperts}, Vicuna-13B-v1.5~\citep{chiang2023vicuna}, Llama2-13B~\citep{touvron2023llama2openfoundation}, and Baichuan2-13B~\citep{yang2023baichuan2openlargescale}. This enables a comprehensive assessment of noise across various dimensions. 

\subsection{Implementation Details}
In our implementation, for similarity computation between queries and documents, we implement the dot product method. We conduct entailment verification using the bart-large-258-mnli-407M model~\cite{lewis2019bartdenoisingsequencetosequencepretraining}, which helps validate the logical relationships between retrieved information and potential answers. Our retrieval corpus consists of the 2018 English Wikipedia dump and current Wikipedia documents, providing a comprehensive knowledge base. Following the challenging setup in previous work~\cite{cuconasu2024power}, we position the ground truth in the middle of the retrieval list rather than at the top. This aims to ensure that our conclusions regarding noise effects more accurately represent real-world scenarios.

\section{Results and Analysis}\label{sec6}
First, we examine the roles of RAG noise (\ref{sec6.1}). While prior work has analyzed its harmful effects, we focus on its beneficial aspects (\ref{sec6.2}). We evaluate these benefits across four dimensions: (1) Generalization across Models, (2) Noise Robustness Across Scenarios, (3) Noise Ratio Impact, and (4) Statistical Validation. Finally, we investigate the underlying mechanisms of these phenomena (\ref{sec6.3}).

\begin{figure}[!ht]
\centering
\includegraphics[width=1.0\linewidth]{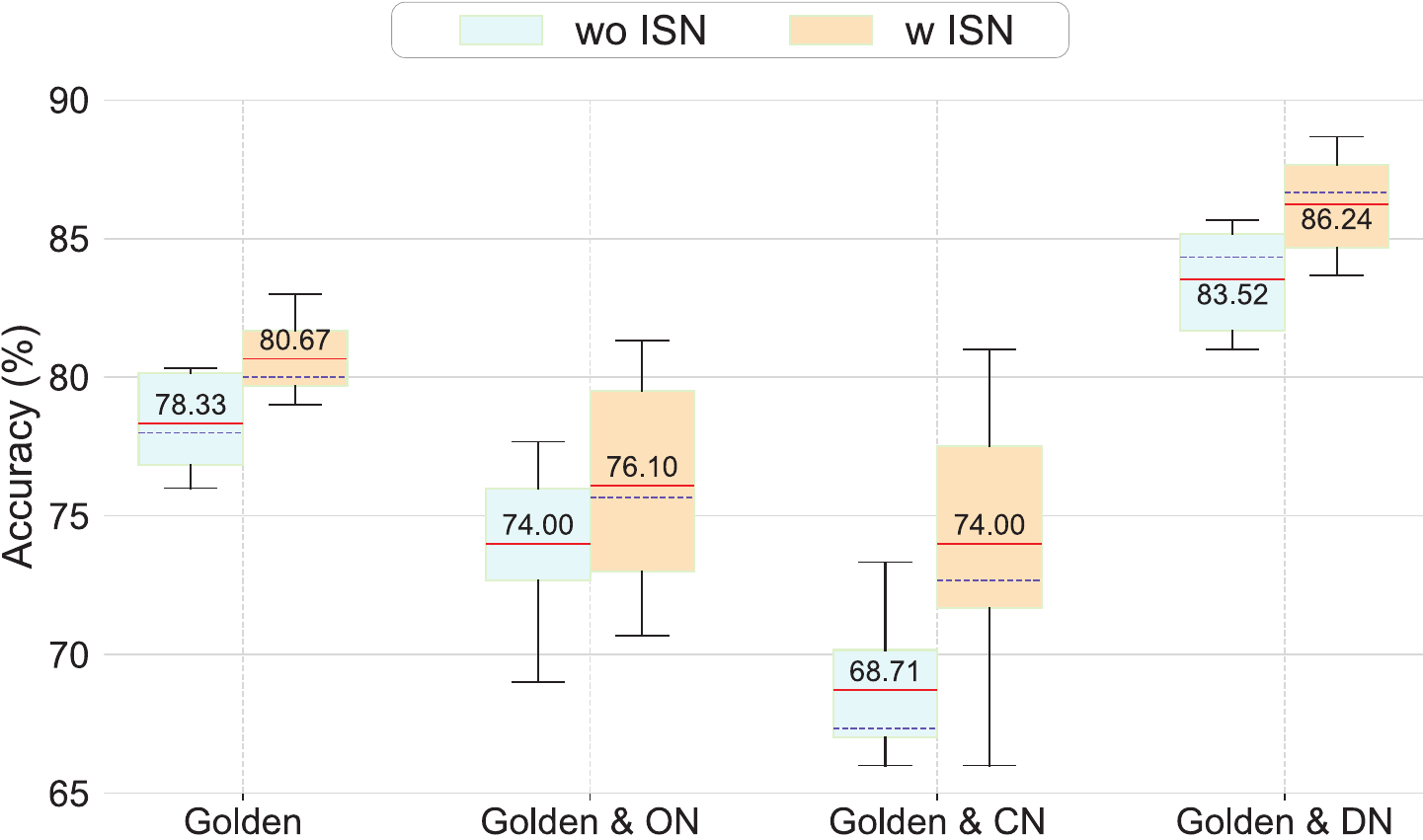}
\caption{Impact of ISN on the average accuracy of eight representative LLMs on RGB. Red solid lines indicate means and purple dashed lines show medians.}
\label{Figure4}
\end{figure}

\subsection{Roles of RAG Noise}\label{sec6.1}
Table \ref{table1} illustrates the impact of diverse noise types (the first six) on two open-source models: Llama3-8B-Instruct and Qwen2-7B-Instruct. We observe consistent performance trends across multiple datasets and retrieval noises. Based on these trends, we can categorize retrieval noises into two types: \textit{harmful noise} (counterfactual, supportive, and orthographic) and \textit{beneficial noise} (semantic, datatype, and illegal sentence). We find that:

(1) For harmful noise, counterfactual noise impacts model performance most significantly by disrupting accurate fact discernment and answer generation. As shown in Figure \ref{Figure1}, the false statement ``Beckham was a prominent player for Manchester United'' leads the model to disregard correct information and respond erroneously. 

\begin{table}[htp]
\centering
\caption{Effects of beneficial noise on Self-RAG (13B). We report enhanced accuracy ratios ($\%$), and the weighted average values (WA, $\%$) are also provided.\label{table2}}
\resizebox{1.0\linewidth}{!}{
\begin{tabular}{l|cccc}
\toprule
{\bfseries Scenario} & {\bfseries NQ} & {\bfseries RGB} & {\bfseries StrategyQA} & {\bfseries WA}\\
\midrule
Golden only & +3.12 & +1.74 & +18.88 & +7.77 \\
Golden \& DN &  +1.84 & +1.96 & +13.50 & +5.49 \\
Golden \& ON & +1.76 & +3.63 & +10.00 & +4.67\\
Average & +2.24 & +2.45 & +14.13 & +5.98 \\
\bottomrule
\end{tabular}}
\end{table}

(2) For beneficial noise, illegal sentence noise exhibits the most notable improvement in model performance. It improves accuracy by an average of 3.32$\%$ and 1.65$\%$ for two models, respectively, and consistently achieves powerful performance across diverse datasets.

\begin{figure*}[ht]
\centering
\includegraphics[width=1.00\linewidth]{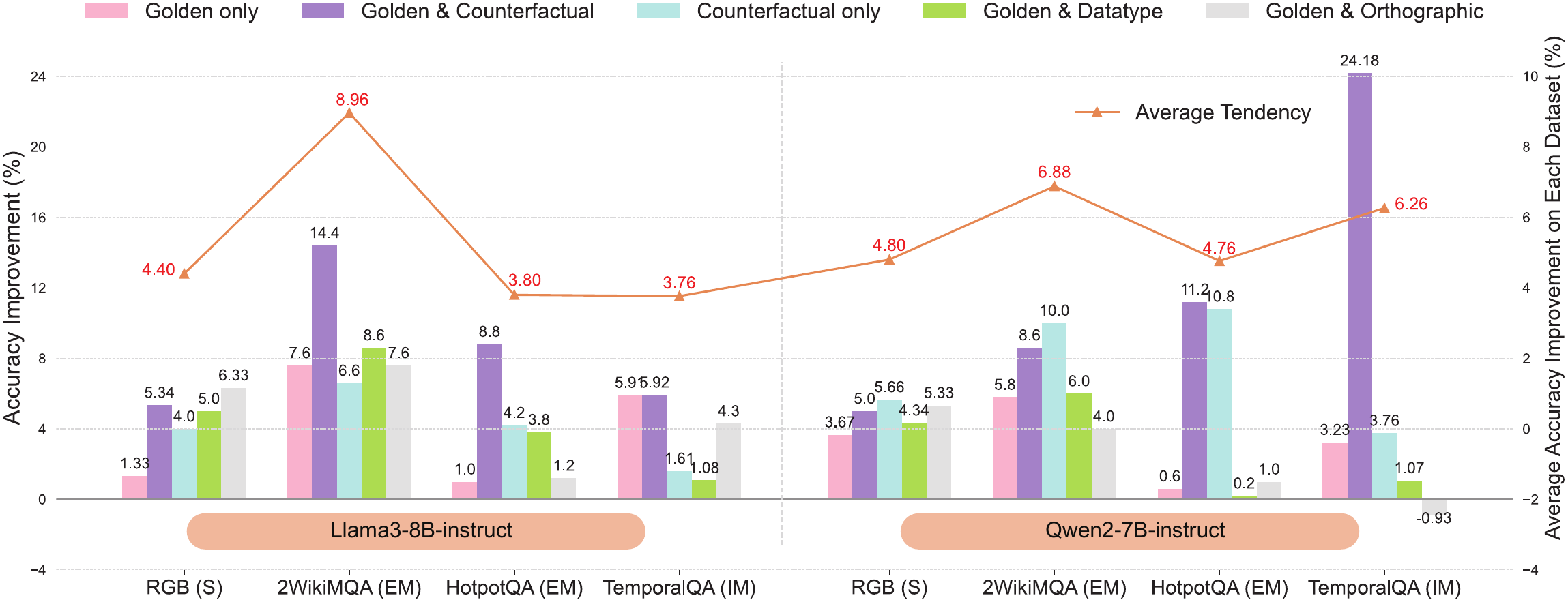}
\caption{Results for the impact of illegal sentence noise on the Llama3-8B-instruct and Qwen2-7B-instruct models when exposed to five typical noise categories across four datasets, including both single-hop (S) and multi-hop (explicit: EM, implicit: IM) reasoning tasks. The bar charts show performance differences upon introducing illegal sentence noise. The line graphs illustrate the average accuracy improvement across noise types per dataset.}
\label{Figure5}
\end{figure*}

\begin{table*}[!ht]
\centering
\caption{Results for different illegal sentence noise (ISN) ratios on RGB. L2-13B, L3-8B, Q2-7B, M-7B, V-13B, B2-13B, L3-70B and M-8x7B represents Llama2-13B, Llama3-8B-Instruct, Qwen2-Instruct, Mistral-7B-Instruct-v0.2, Vicuna-13B-v1.5, Baichuan2-13B-chat, Llama3-70B-Instruct, Mixtral-8x7B-Instruct.}
\label{table3}
\resizebox{1\textwidth}{!}{
\begin{tabular}{l|cccccc|cc|l}
\toprule
\multirow{2.5}{*}{{\bfseries Scenario}} & \multicolumn{6}{c|}{{\bfseries Small}} & \multicolumn{2}{c|}{{\bfseries Large}} & \multirow{2}{*}{{\bfseries Average}} \\
\cmidrule(lr){2-7} \cmidrule(lr){8-9}
\quad           & L2-13B & L3-8B & Q2-7B & M-7B & V-13B & B2-13B & L3-70B & M-8x7B & \\
\midrule
0               & 29.33  & 80.00 & 76.33 & 80.33 & 80.33 & 78.00 & 76.00 & 77.33 & 72.21 \\
\quad + ISN     & 72.33  & 83.00 & 80.00 & 81.00 & 82.33 & 79.67 & 79.67 & 79.67 & {\bfseries 79.71}$_{\textcolor{red}{+7.50}}$ \\
\midrule
0.2             & 18.67  & 77.33 & 75.33 & 76.00 & 79.33 & 73.33 & 76.67 & 73.67 & 68.79 \\
\quad + ISN     & 73.67  & 82.67 & 80.33 & 76.67 & 80.00 & 72.33 & 80.33 & 73.67 & {\bfseries 77.46}$_{\textcolor{red}{+8.67}}$ \\
\midrule
0.4             & 12.33  & 73.67 & 71.33 & 69.00 & 72.67 & 68.00 & 76.33 & 65.67 & 63.63 \\
\quad + ISN     & 70.67  & 77.00 & 73.00 & 71.00 & 73.33 & 68.33 & 80.00 & 66.67 & {\bfseries 72.50}$_{\textcolor{red}{+8.87}}$ \\
\bottomrule
\end{tabular}}

\end{table*}

For prior noise, we evaluate on our PriorQA dataset in Appendix Table \ref{table7}. Questions in PriorQA contain factual errors, such as ``Which country hosted 1980 FIFA World Cup?'' (1980 FIFA World Cup was not held). Accuracy is measured by whether LLMs correctly identify and respond with ``The question is factually incorrect''. LLMs achieve 79.93$\%$ average accuracy in handling prior noise. However, when models fail to identify prior errors and continue retrieval, accuracy drops to 34.20$\%$. This highlights the importance of detecting factual errors in queries before generating responses.

\subsection{Additional Results on Beneficial Noise}\label{sec6.2}

\noindent
\textbf{Generalization across Models}\; 
To demonstrate beneficial noise's broad applicability, we examine its effects across model architectures (Figure \ref{Figure4}) and RAG configurations (Table \ref{table2}). For brevity, we present illegal sentence noise results in the main text, with full results in the Appendix.

Results across various architectures and scales are shown in Figure \ref{Figure4}, we evaluate the impact of illegal sentence noise (ISN) on eight LLMs by presenting average accuracy across scenarios with no noise, harmful noise (e.g. CN, ON), and beneficial noise (e.g. DN). We apply proportional scaling to CN data to make a clearer illustration within one figure while maintaining consistent conclusions. The results indicate that ISN significantly enhances model performance in all scenarios, with the most substantial improvement under harmful noise.

Noise effects on specialized RAG models are illustrated in Table \ref{table2}. Introducing illegal sentence noise to the specialized RAG model Self-RAG~\citep{asai2024selfrag} consistently enhances model performance across various datasets (NQ, RGB, and StrategyQA) and scenarios (without noise, with harmful or beneficial noise). This further validates positive effects of beneficial noise.

\noindent\textbf{Noise Robustness Across Scenarios}\; 
We analyze the effect of illegal sentence noise (ISN) in 5 scenarios: no noise (i.e., Golden only), harmful noise (i.e., Golden $\&$ Counterfactual, Counterfactual only and Golden $\&$ Orthographic), and beneficial noise (i.e., Golden $\&$ Datatype). Figure \ref{Figure5} shows accuracy gains with ISN introduction. Results indicate consistent improvements across datasets, especially when combined with harmful noise like counterfactual, leading to an average accuracy increase of over 10$\%$. This highlights the potential significance of beneficial noise in RAG applications.

\begin{table}[!htp]
\centering
\caption{Statistical significance of differences between scenarios with and without beneficial noises.\label{table4}}
\resizebox{1.0\linewidth}{!}{
\begin{tabular}{c|cc}
\toprule
{\bfseries Noise} & {\bfseries Llama3-8B-Instruct} & {\bfseries Qwen2-7B-Instruct}\\
\midrule
ISN & 4.10e-5 & 4.88e-3 \\
DN & 1.71e-4 & 9.59e-4 \\
\bottomrule
\end{tabular}}
\end{table}

\vspace{0.5em}\noindent
\textbf{Noise Ratio Impact}\; 
To demonstrate the positive effects at different harmful noise ratios, we present results for orthographic noise disturbances with ratios ranging from 0 to 0.4. As shown in Table \ref{table3}, we see that the introduction of illegal sentence noise (beneficial noise) consistently enhances model performance, thereby further illustrating the generalizability of beneficial noise.

\vspace{0.5em}\noindent
\textbf{Statistical Validation}\; 
To statistically evaluate the differences between scenarios with and without beneficial noise, we apply the nonparametric Wilcoxon signed-rank test~\citep{wilcoxon1992individual}. This method effectively measures the magnitudes of differences and detects statistical significance between two conditions. We test the null hypothesis of no significant difference ($H_0: difference = 0$) against the alternative hypothesis of a significant difference ($H_1: difference \neq 0$). Following~\citep{seth2023comparing, wu2023kgetcda}, we use a significance level of 0.05. As shown in Table \ref{table4}, all p-values are below 0.05, leading us to reject the null hypothesis ($H_0$). These results provide strong statistical evidence that beneficial noise improves model performance.

\subsection{Analysis of Noise Phenomena}\label{sec6.3}
We propose three hypotheses regarding how beneficial noise may enhance performance, which we confirm through case study and statistical analysis.

\begin{itemize}
    \item \textit{{\bfseries H1: Clearer reasoning process}}
    \item \textit{{\bfseries H2: More standardized response formats}} 
    \item \textit{{\bfseries H3: Increased confidence with gold context}}
\end{itemize}

\begin{figure}[ht]
\centering
\includegraphics[width=1.00\linewidth]{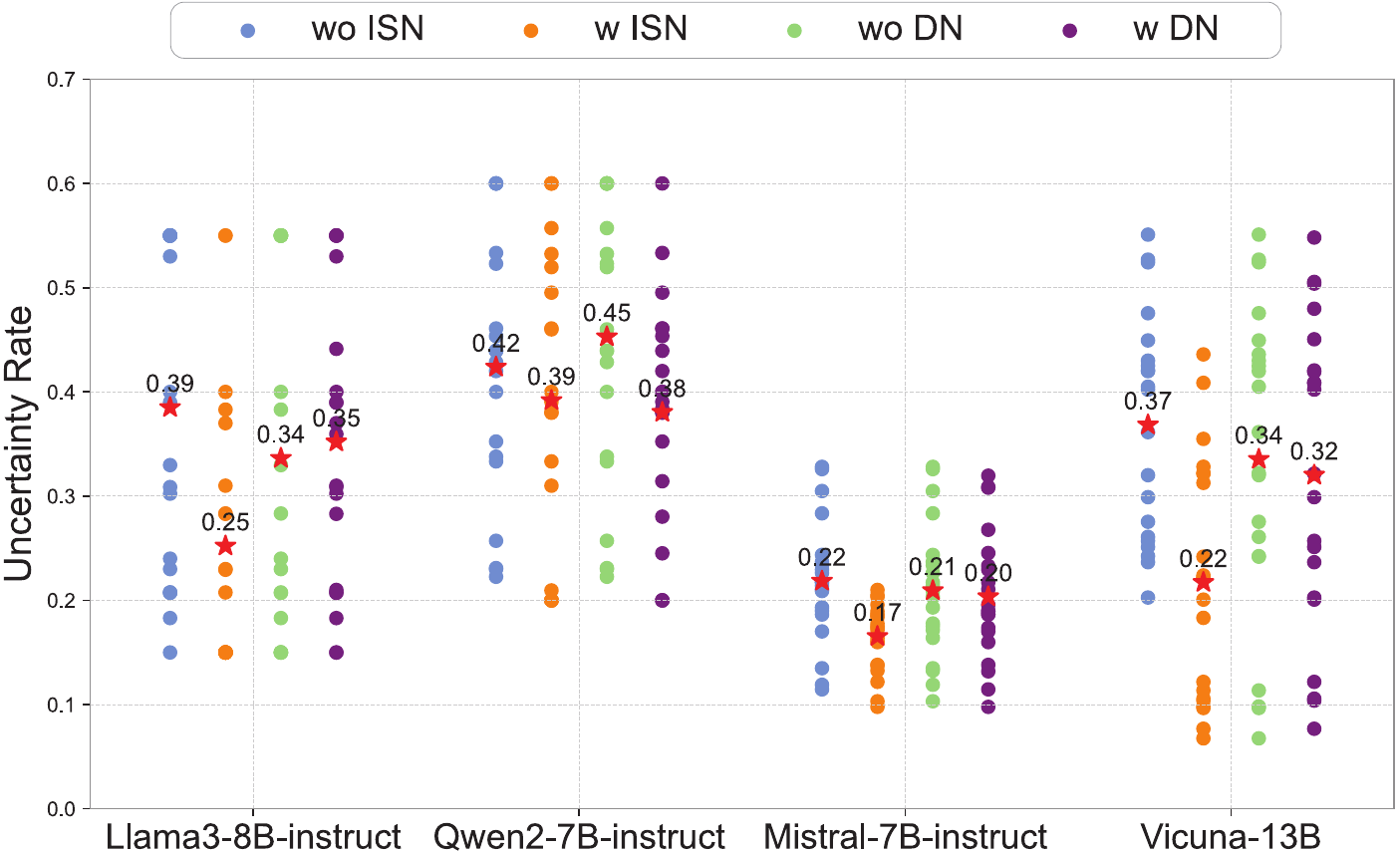}
\caption{Impact of beneficial noise on LLM output uncertainty (anti-confidence). ISN and DN represent Illegal Sentence Noise and Datatype Noise, with \textcolor{red}{$\star$} indicating mean uncertainty rate (µ). Results show that LLMs pay more attention to the provided golden context and respond with greater confidence.}
\label{Figure6}
\end{figure}

\vspace{0.5em}\noindent
\textbf{Illustrative Case Study}\; 
Table \ref{table13} in the appendix presents the reasoning process of Llama3-8B-instruct on the multi-hop dataset Bamboogle. Without beneficial noise, the model ignores correct information and exhibits logical flaws under counterfactual noise influence. This is exemplified by its erroneous statement: ``The other options are incorrect, as they provide different birth dates for the author.'' However, upon introducing beneficial noise, the model exhibits heightened attention to the golden context and successfully distinguishes between correct and incorrect information (\textit{{\bfseries H1}}). We hypothesize that beneficial noise enhances the LLMs' ability to integrate its parameterized knowledge with retrieved information, thus improving its capacity to discern truth from falsehood. Furthermore, by comparing model outputs under two conditions, we observe that beneficial noise contributes to more standardized answer formats (\textit{{\bfseries H2}}).

\vspace{0.5em}\noindent
\textbf{Statistical Characterization}\;
To verify three hypotheses statistically, we use a two-step process. We first gather model outputs from multiple datasets before and after introducing beneficial noise. Then, we randomly sample 100 examples per dataset to manually assess which condition produces more standardized output formats and clearer reasoning processes. Outputs are deemed similar if no significant difference exists between conditions with and without beneficial noise.
Results across seven datasets show that, on average, 37 samples with beneficial noise exhibit clearer reasoning compared to 31 without (\textit{{\bfseries H1}}), while 26 samples with beneficial noise demonstrate better output formats versus 23 without (\textit{{\bfseries H2}}).

Second, as shown in Figure \ref{Figure6}, we analyze the impact of beneficial noise on LLM output uncertainty across four powerful LLMs. Results indicate that when combined with beneficial noise (ISN or DN), LLMs generally exhibit lower uncertainty and increased confidence in their outputs. This suggests that LLMs pay more attention to provided golden context and respond with greater confidence (\textit{{\bfseries H3}}).

\section{Conclusion}\label{sec7}
We define and categorize seven types of RAG noise into beneficial and harmful groups, exploring retrieval noise from linguistic and practical perspectives. To conduct this evaluation, we propose a systematic framework for generating various retrieval documents and establish a novel noise benchmark, NoiserBench. Our experiments reveal that beneficial noise can significantly enhance model performance through clearer reasoning paths, standardized answers, and increased confidence—acting much like Aladdin's Lamp. These findings may offer insights for leveraging beneficial noise mechanisms in future research.

\section*{Limitations}
While our systematic analysis of RAG noises in real-world scenarios offers valuable insights, several limitations warrant consideration. First, our analysis of noise phenomena remains relatively preliminary. Future work will examine the underlying mechanisms by investigating parameter variations, particularly attention values, across each model layer. In addition, future work could explore the effects of noise across a wider variety of task domains, including complex reasoning, where noise may interfere with multi-step inference and lead to compounding errors. Expanding the scope in this direction could help develop more robust retrieval-augmented generation systems for real-world applications.

\section*{Acknowledgement}
We thank the anonymous reviewers for their feedback on this work. This work is supported by the National Key R$\&$D Program of China under Grant No.2024YFB2808802 and Postdoctoral Fellowship Program of CPSF (GZC20240840).

\bibliography{custom}

\appendix

\section*{Appendix}
\label{sec:appendix}
Within this supplementary material, we elaborate on the following aspects:
\begin{itemize}
    \item Appendix A: Models
    \item Appendix B: Implementation Details
    \item Appendix C: Results
\end{itemize}

\section{Models}
We provide brief introductions to LLMs used in our experiments. For more details, please refer to the official websites or the corresponding Hugging Face Transformers repository.

\begin{itemize}
    \item \textbf{Llama2 $\&$ Llama3}: The Llama series model, developed by Meta AI's FAIR team, is a widely-used autoregressive language model. These models, particularly Llama3, achieve competitive performance compared to some state-of-the-art closed-source LLMs. We use the 13B model for Llama2, and the 8B and 70B models for Llama3.
    
    \item \textbf{Vicuna-v1.5}: The Vicuna model, derived from fine-tuning the LLaMA-2 base model by LMSYS, was developed using around 70K user-shared conversations obtained from ShareGPT.com through public APIs. We use the popular vicuna-13B here.

    \item \textbf{Qwen2}: Proposed by Alibaba Cloud, Qwen series are strong language models, which have been stably pretrained for up to 3 trillion tokens of multilingual data with a wide coverage of domains, languages (with a focus on Chinese and English), etc. Qwen2-7B-Instruct is utilized.

    \item \textbf{Mistral}: The Mistral series includes the Mistral-7B and Mixtral-8x7B models. The Mistral-7B is an autoregressive language model with 7 billion parameters, trained on a diverse corpus to ensure high performance in various tasks. The Mixtral-8x7B is a high-quality sparse mixture of expert models (SMoE) with open weights. This technique increases the number of parameters of a model while controlling cost and latency, as the model only uses a fraction of the total set of parameters per token.

    \item \textbf{Baichuan2}: Baichuan2 is the new generation of open-source language models launched by Baichuan Intelligence. It is trained on a high-quality corpus with 2.6 trillion tokens and has achieved the best performance in authoritative Chinese and English benchmarks of the same size. We use the 13B chat model.
\end{itemize}

\section{Implementation Details}
\subsection{Compute Infrastructure}
We execute the experiments using the following compute specifications.
\begin{itemize}
    \item NVIDIA A100 80 GB GPU $\times$ 2
    \item 256 GB RAM
\end{itemize}

We use Python 3.10.0 and speed up inference using vllm\footnote{\url{https://github.com/vllm-project/vllm}}, a fast and easy-to-use library. In Table \ref{table5}, we list the main libraries along with their versions.

\begin{table}[!ht]
\centering
\caption{Main libraries and the corresponding versions.\label{table5}}
\begin{tabular}{p{2.5cm}|p{2.5cm}}
\toprule
{\bfseries Package} & {\bfseries Version}\\
\midrule
vllm & 0.2.6 \\
torch & 2.1.2+cuda12.4 \\
transformers & 4.36.2\\
\bottomrule
\end{tabular}
\end{table}

\begin{figure}[!htp]
\centering
\includegraphics[width=0.97\linewidth]{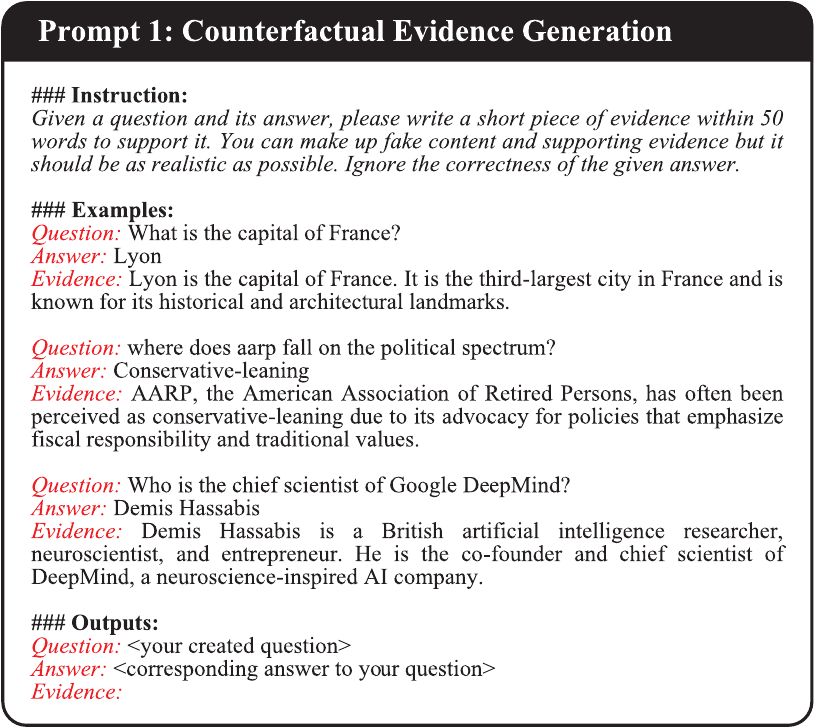}
\caption{Example LLMs' input for counterfactual evidence generation. This prompt is composed of instruction, examples, and candidate counterfactual QA.}
\label{FigureS10}
\end{figure}

\subsection{Dataset Construction}
To construct our benchmark NoiserBench, we need to first gather candidate QA instances from multiple sources. In this paper, our source data is obtained from seven publicly available datasets, including single-hop NQ and RGB, explicit multi-hop HotpotQA, 2WikiMQA, Bamboogle, and implicit multi-hop StrategyQA and TempQA. Table \ref{table6} shows the full list of candidate instances, and in total, we use 26,855 instances.

\begin{figure}[htp]
\centering
\includegraphics[width=0.97\linewidth]{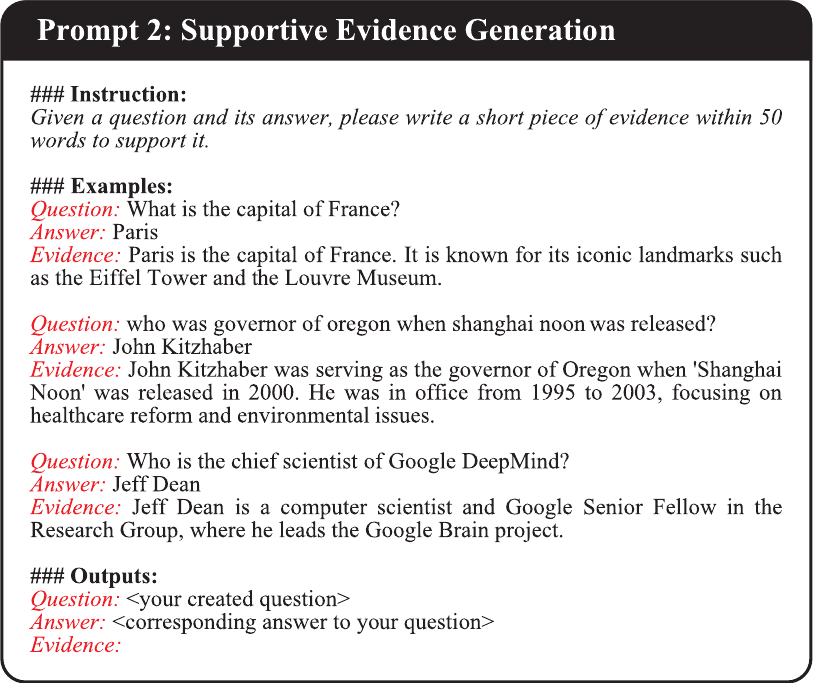}
\caption{Example LLMs' input for supportive evidence generation. This prompt is composed of instruction, examples, and candidate QA.}
\label{FigureS11}
\end{figure}

\begin{figure}[ht]
\centering
\includegraphics[width=0.97\linewidth]{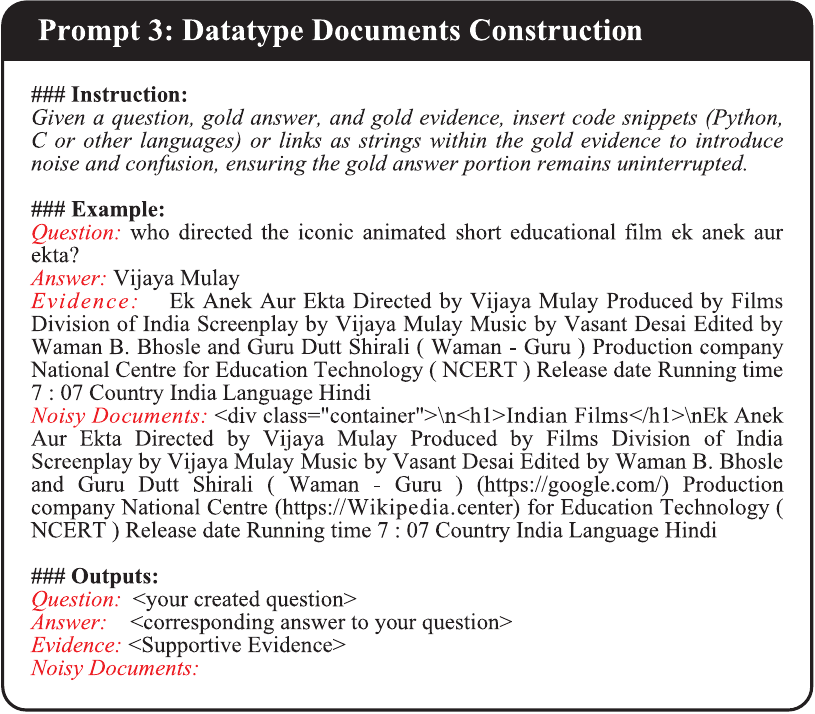}
\caption{Example LLMs' input for datatype noise construction. This prompt is composed of instruction, examples, candidate QA and corresponding evidence.}
\label{FigureS12}
\end{figure}

Subsequently, we introduce various noisy documents using external tools. For counterfactual noise, we obtain relevant entities related to the golden answer from Google search\footnote{We query Google search via SerpAPI: \url{https://serpapi.com}} to construct counterfactual answers. For orthographic noise, we utilize the open-source textnoisr package\footnote{\url{https://github.com/preligens-lab/textnoisr}}, which enables the convenient introduction of noise to text datasets and precise control of the quality of results. Four types of ``action'' are implemented: insert, delete, substitute, and swap. For other types of noise, we utilize the 2018 English Wikipedia dump for document construction. We present the prompts in Figure \ref{FigureS10}-\ref{FigureS12}.

\subsection{Additional Details}
We utilize a CN for retrieval, where relevant Chinese documents are retrieved in response to the input query to enhance the prompt, rather than relying on a fixed set of N Chinese examples. This CN corpus is constructed using data from Wikidata and Google Search, intentionally designed to include misleading or outdated information in order to simulate real-world scenarios where context might be inaccurate or evolving.

During the experiment, we found that the optimal Top-k value for the contriver was 5, and the similarity threshold for all-MiniLM-L6-v2 was set to 0.3.

\begin{table*}[!ht]
\centering
\caption{Statistics of source QA instances from a couple of knowledge-intensive datasets. `E' and `I' represent explicit and implicit, respectively.}
\label{table6}
\small 
\renewcommand{\arraystretch}{1.2} 

\resizebox{1\textwidth}{!}{
\begin{tabularx}{\textwidth}{l|c|c|c|c|X}
\toprule
{\bfseries Dataset} & {\bfseries Category} & {\bfseries Source} & {\bfseries \#Source pairs}&{\bfseries \#Samples} & \multicolumn{1}{c}{\bfseries Example}\\
\midrule
NQ & Single-hop & Train set & 2,889&500 & Who won the 7 man elimination chamber match?\\
RGB & Single-hop & Test set & 300&300 & How many vehicles did Tesla deliver in 2021?\\
HotpotQA & Multi-hop (E) & Dev set & 7,405&500 & What election will take place on the same day as the United States Senate election in Texas?\\
2WikiMQA & Multi-hop (E) & Dev set & 12,576&500 & Where was the place of death of Isabella of Bourbon's father?\\
Bamboogle & Multi-hop (E) & All & 125&125 & Who was the first African American mayor of the most populous city in the United States?\\
StrategyQA & Multi-hop (I) & Train set & 2,290&500 & Can Arnold Schwarzenegger deadlift an adult Black rhinoceros?\\
TempQA & Multi-hop (I) & All & 1,270&186 & Who was the commander-in-chief of the colonial army during the revolutionary war?\\
PriorQA & Mix-hop & All & 500&500 & What were the primary strategies employed by the British army duringthe American Civil War?\\
\bottomrule
\end{tabularx}
}
\end{table*}

\section{Results}
In this section, we provide supplementary results to further illustrate the role of RAG noise, especially beneficial noise. Our analysis primarily focuses on datatype noise, orthographic noise, and prior noise, as illegal sentence noise has been extensively discussed in the main text, and other forms of noise have been explored in previous studies. These additional results aim to provide a more comprehensive understanding of various noise types and their effects on the model's performance.

\begin{table}[!h]
\centering
\caption{The effects of prior noise on LLMs, which is measured by accuracy (\%). `Base' indicates the scenario with no retrieval. `Misleading' refers to counterfactual content associated with prior noise. `Background' denotes multiple retrieval results obtained after decomposing the query into its constituent entities.\label{table7}}
\small 
\renewcommand{\arraystretch}{1.1} 
\resizebox{\columnwidth}{!}{ 
\begin{tabularx}{\columnwidth}{l|>{\centering\arraybackslash}X>{\centering\arraybackslash}X>{\centering\arraybackslash}X}
\toprule
{\bfseries Models} & {\bfseries Base} & {\bfseries Misleading} & {\bfseries Background}\\
\midrule
Llama3-8B      & 93.40 & 47.80 & 90.00 \\
Qwen2-7B       & 94.20 & 28.20 & 98.20 \\
Mistral-7B     & 96.60 & 28.60 & 99.20 \\
Llama2-13B     & 21.00 & 5.60  & 61.60 \\
Vicuna-13B     & 91.00 & 25.80 & 99.20 \\
Baichuan2-13B  & 90.00 & 45.20 & 96.40 \\
Llama3-70B     & 99.00 & 78.40 & 99.80 \\
Mixtral-8x7B   & 91.20 & 39.00 & 99.60 \\
\midrule
{\bfseries Average} & {\bfseries 79.93} & {\bfseries 34.20} & {\bfseries 88.47} \\
\bottomrule
\end{tabularx}
}
\end{table}

\begin{table*}[ht]
\centering
\caption{Impact of various noise types on accuracy (\%) for eight representative LLMs on the RGB dataset. We assess performance across various retrieval scenarios: ``Base'' (no retrieval), ``Golden Only'' (only golden retrieval context), and ``Golden \& XXX'' (golden context + specific retrieval noises, including Counterfactual, Supportive, Orthographic, Semantic, Datatype, Illegal Sentence Noise).}\label{table8}
\resizebox{1\textwidth}{!}{
\begin{tabular}{l|cccccc|cc|c}
\toprule
\multirow{2}{*}{{\bfseries Scenario}} & \multicolumn{6}{c|}{{\bfseries Small}} & \multicolumn{2}{c|}{{\bfseries Large}} & \multirow{2}{*}{{\bfseries Average}} \\
\cmidrule(lr){2-7} \cmidrule(lr){8-9}
 & L2-13B & L3-8B & Q2-7B & M-7B & V-13B & B2-13B & L3-70B & M-8x7B & \\
\midrule
Base          & 17.00  & 47.00 & 31.33 & 27.00 & 35.33 & 27.67  & 60.00  & 43.00  & 36.04 \\
Golden Only   & 29.33  & 80.00 & 76.33 & 80.33 & 80.33 & 78.00  & 76.00  & 77.33  & 72.20 \\
Golden \& CN  & 14.00  & 36.33 & 26.00 & 19.33 & 19.33 & 15.00  & 42.33  & 31.00  & 25.42 \\
Golden \& SuN & 26.00  & 80.00 & 74.00 & 72.33 & 61.67 & 65.33  & 73.67  & 76.67  & 66.21 \\
Golden \& ON  & 14.33  & 75.00 & 74.00 & 72.67 & 77.67 & 69.00  & 77.00  & 72.67  & 66.54 \\
Golden \& SeN & 18.00  & 81.33 & 77.67 & 56.67 & 52.00 & 59.00  & 76.33  & 77.33  & 62.30 \\
Golden \& DN  & 40.00  & 81.67 & 84.33 & 85.33 & 85.67 & 81.67  & 85.00  & 81.00  & 78.08 \\
Golden \& ISN & 72.33  & 83.00 & 80.00 & 81.00 & 82.33 & 79.67  & 79.67  & 79.00  & 79.63 \\
\bottomrule
\end{tabular}}
\end{table*}

\subsection{Results on Prior Noise}
Table \ref{table7} presents results for RAG models affected by prior noise using our dataset, PriorQA. Questions in this dataset contain factual errors, such as ``Which country hosted the 1980 FIFA World Cup?'' (Actually, 1980 FIFA World Cup was not held). Accuracy is assessed by whether models correctly identify and respond with ``The question is factually incorrect''. We observe that all models except Llama2-13B perform well with direct prompts and benefit from retrieving background information due to extensive pre-training knowledge. However, models like Llama2-13B, which persist in searching based on incorrect priors, may retrieve false information and exhibit diminished performance. This underscores the need to detect prior errors in user queries before answering in future RAG system designs.

\begin{figure}[ht]
\centering
\includegraphics[width=1.00\linewidth]{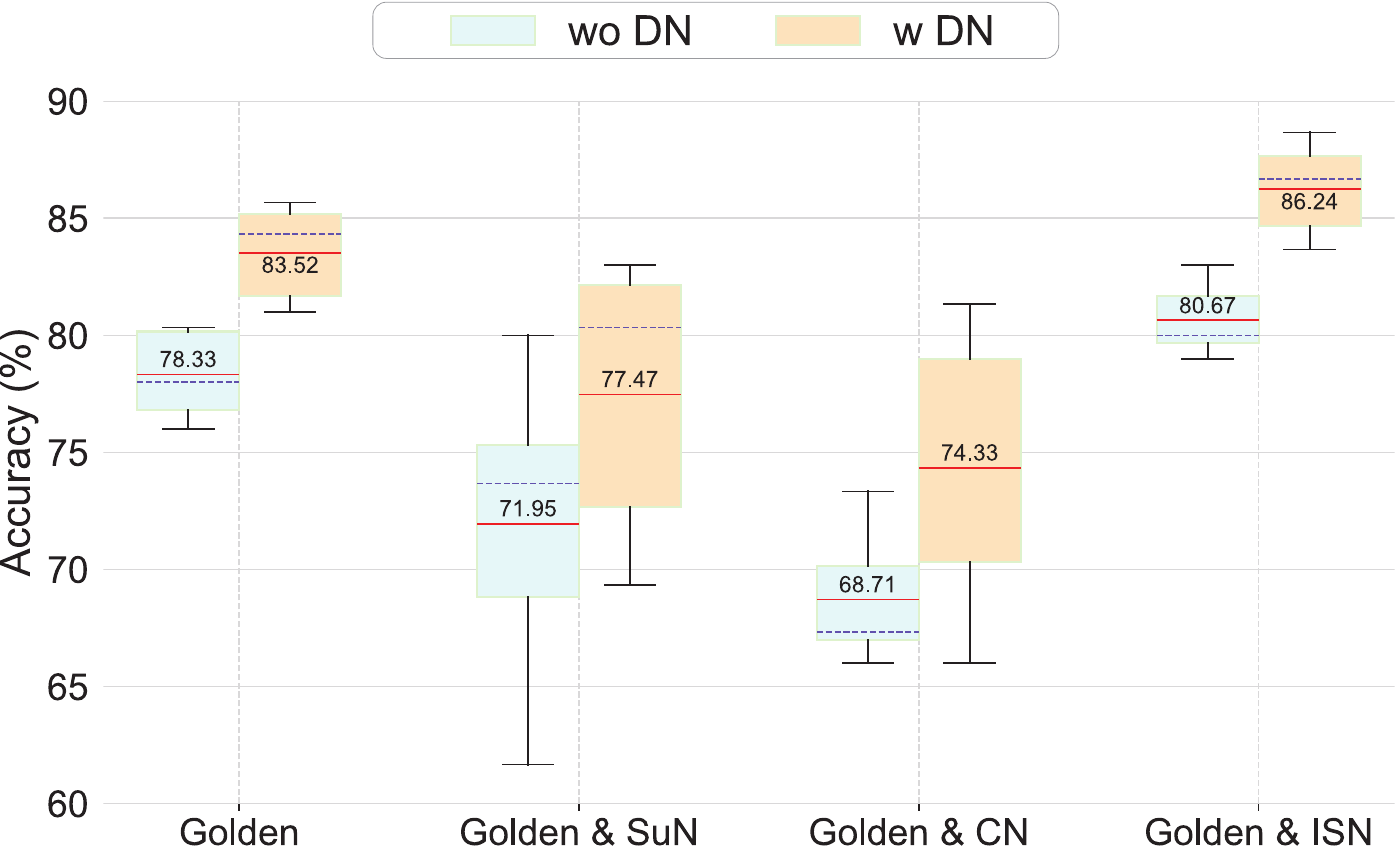}
\caption{Impact of datatype noise (DN) on the average accuracy of eight representative LLMs on RGB. `Golden', `SuN', `CN', and `ISN' represent golden context only, golden context with supportive, counterfactual, and illegal sentence noise, respectively. The mean is marked by a red solid line and the median by a purple dashed line.}
\label{FigureS7}
\end{figure}

\subsection{Results Across Eight Models}
As shown in Figure \ref{FigureS7}, we first present the average performance over seven datasets for datatype noise to demonstrate that beneficial noise improves performance across various LLMs with different model architectures and scales. We apply proportional scaling to counterfactual data to make a clearer illustration within one figure while maintaining consistent conclusions. The results indicate that datatype noise significantly enhances model performance in all scenarios, with the most substantial improvement under harmful noise.

Additionally, we provide detailed results for eight models on the RGB dataset, which is based on recent news corpora and thus better reflects the impact of noise. As shown in Table \ref{table8}, we have the following three findings:

\begin{figure*}[ht]
\centering
\includegraphics[width=1.00\linewidth]{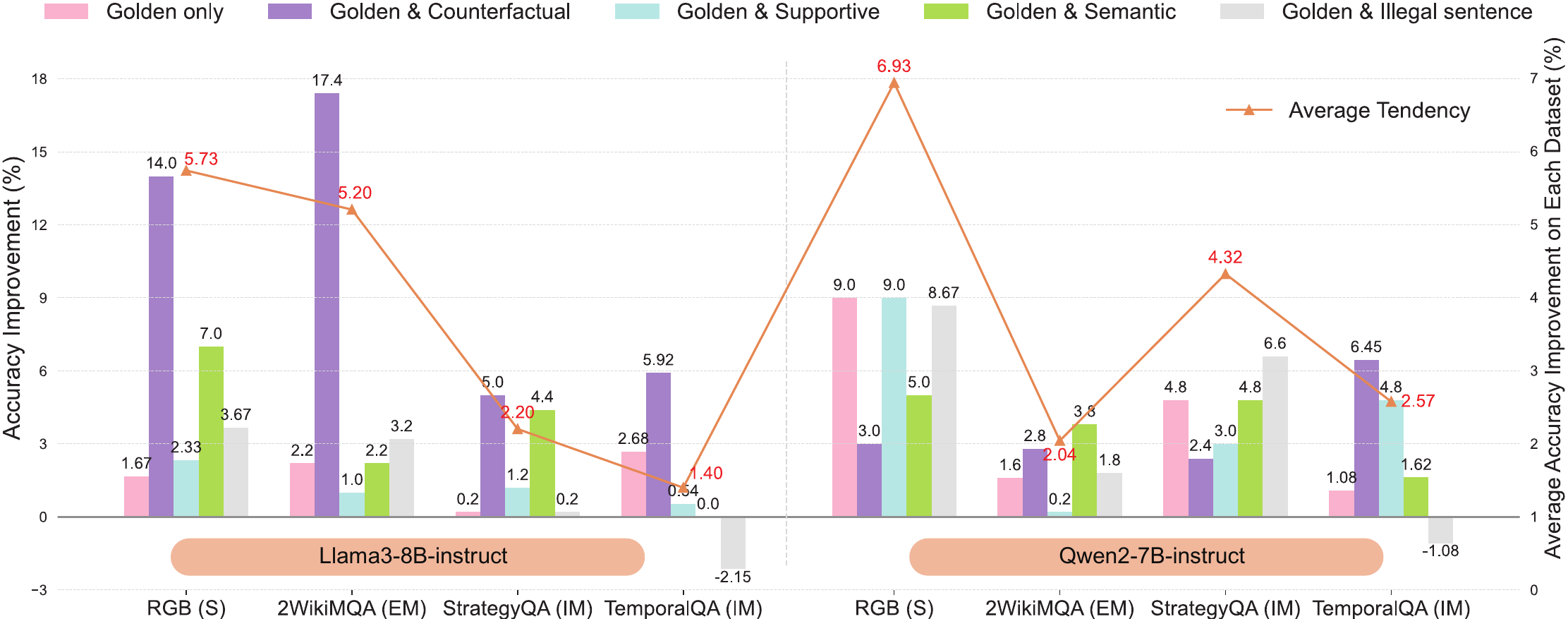}
\caption{Results for the impact of datatype noise on the Llama3-8B-instruct and Qwen2-7B-instruct models when exposed to five typical noise categories across four datasets, including both single-hop (S) and multi-hop (explicit: EM, implicit: IM) reasoning tasks. The bar charts show performance differences upon introducing datatype noise. The line graphs illustrate the average accuracy improvement across noise types per dataset.}
\label{FigureS8}
\end{figure*}

\begin{itemize}[leftmargin=10pt]
    \item \textbf{Global Impact of Beneficial Noise (DN, ISN):} Datatype Noise (DN) and Illegal Sentence Noise (ISN) consistently improve performance across all model scales and capabilities, with average improvements of 5.8$\%$ and 7.4$\%$ respectively over the golden-only baseline. This demonstrates the universal applicability of these beneficial noise types.
    \item \textbf{Global Impact of Harmful Noise (CN, SuN, ON):} Counterfactual Noise (CN), Supportive Noise (SuN), and Orthographic Noise (ON) consistently degrade performance across all models, with CN showing the most severe negative impact (average performance decrease of 46.8$\%$ compared to the golden-only baseline).
     \item \textbf{Scale-Dependent Semantic Noise Effects:} The impact of Semantic Noise (SeN) is twofold. For \textbf{less optimized models} (e.g., Llama2-13B, Mistal-7B), SeN acts as harmful noise. This may be due to smaller models being less confident in their parametric memory and having weaker reasoning capabilities. Consequently, they are more easily misled by semantically irrelevant context, consistent with findings from previous studies~\cite{pmlr-v202-shi23a,xie2024adaptive}; For \textbf{larger models} (e.g., Llama3-70B), SeN becomes beneficial. Larger models, with more robust parametric memory and better understanding, are less susceptible to irrelevant context. They can efficiently ignore semantically unrelated content and focus on core details necessary to answer questions, leading to performance gains.
\end{itemize}

\subsection{Results Across the Number of Quires}
We leverage keywords extracted from queries (ranging from 1 to 4 keywords per query) for content retrieval. As shown in Table~\ref{table10}, results on 2WikiMQA demonstrate that the core findings regarding the impact of beneficial and harmful noise remain consistent across queries of varying complexity. 

\begin{table}[!htp]
\centering
\caption{Accuracy in different scenarios and with different numbers of queries.}\label{table10}
\resizebox{0.4\textwidth}{!}{
\begin{tabular}{l|cccc}
\toprule
\multirow{2.5}{*}{{\bfseries Scenario}} & \multicolumn{4}{c}{{\bfseries Number of queries}}  \\
\cmidrule(lr){2-5} 
 & 1 & 2 & 3 & 4 \\
\midrule
G only          & 79.80  & 84.20 & 84.60 & 84.60  \\
G\&CN   & 21.20  & 23.60 & 23.60 & 23.80  \\
G\&ISN    & 87.40  & 89.40 & 89.20 & 89.40 \\
\bottomrule
\end{tabular}}
\end{table}

\begin{figure}[!ht]
\centering
\includegraphics[width=1.00\linewidth]{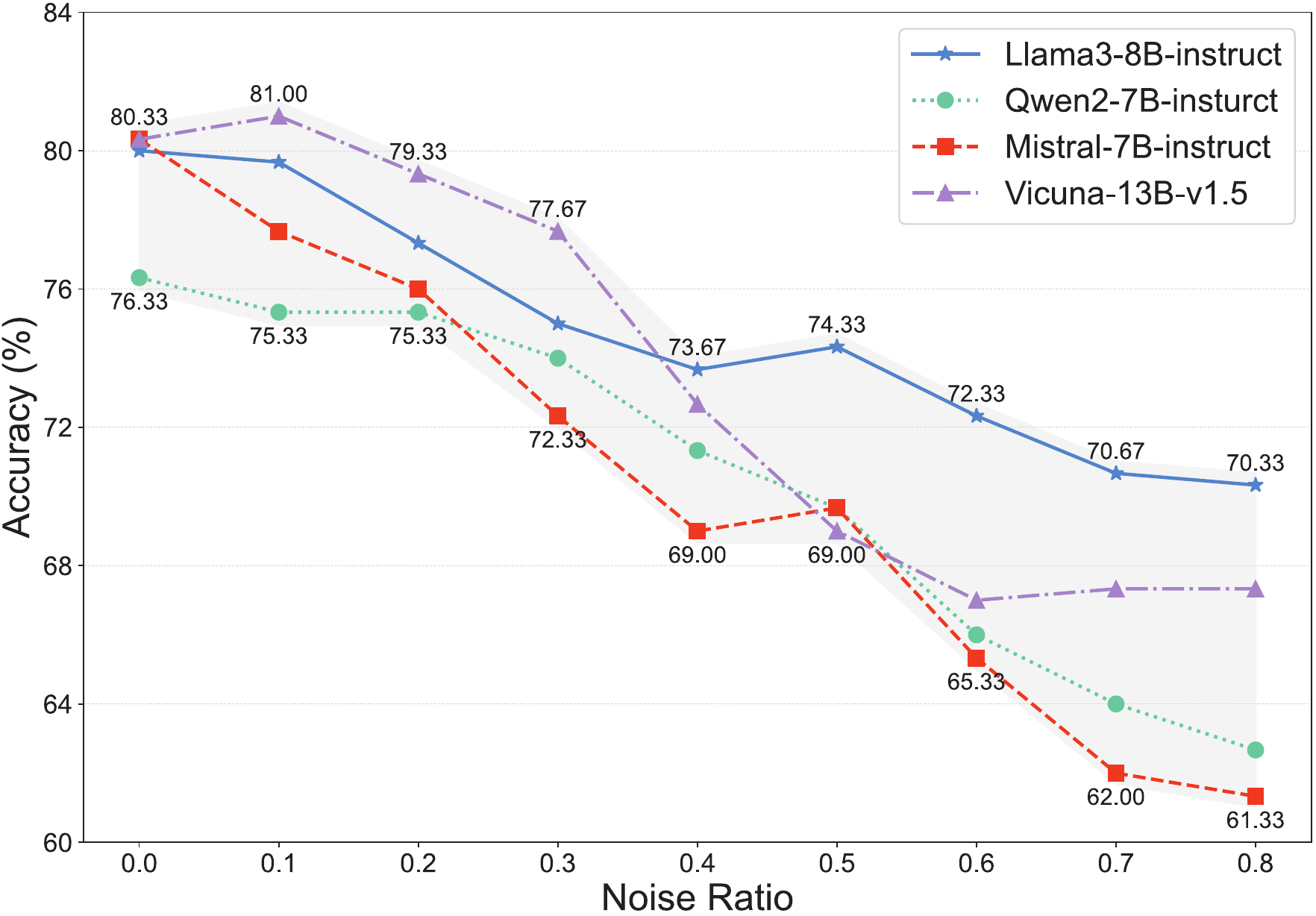}
\caption{The experimental results of noise robustness measured by accuracy ($\%$), under different orthographic noise ratios. Performance is benchmarked across state-of-the-art open-source models, such as Llama3-8B-instruct, for noise ratios ranging from 0 to 0.8. The maximum and minimum accuracy for all models at each noise ratio is annotated, with a shaded region representing ±0.5$\%$ threshold to illustrate the overall trend in model performance better as the noise ratio increases. }
\vskip -0.1in
\label{FigureS9}
\end{figure}

\subsection{Performance Under Other Noise Disturbances}
To illustrate the impact of beneficial noise under other noise disturbances, we analyze the effect of datatype noise (DN) in five scenarios: no noise (i.e., Golden only), harmful noise (i.e., Golden $\&$ counterfactual noise, Golden $\&$ supportive noise), and beneficial noise (i.e., Golden $\&$ illegal sentence noise, Golden $\&$ Semantic noise). Figure \ref{FigureS8} shows the model's accuracy gains after introducing DN in these scenarios. We find that DN generally enhances performance across all datasets, particularly when combined with harmful noise like counterfactual noise, with average accuracy improvements exceeding 10 percentage points. This consistent enhancement underscores beneficial noise's potential significance for future RAG research.

\subsection{Noise Robustness of RAG Models under Different Noise Ratios}
We provide the results of four representative LLMs under different orthographic noise ratios. Specifically, for insert, delete, and substitute actions, the noise ratio ranges from 0.0 to 0.9, while for swapping, it ranges from 0.0 to a maximum of 0.5. As shown in Figure \ref{FigureS9}, the maximum and minimum accuracy for all models at each noise ratio is annotated, with a shaded region representing ±0.5$\%$ threshold to better illustrate the overall trend in model performance as the noise ratio increases. We observe that increasing noise rates pose a challenge for RAG in LLMs, particularly when the ratio exceeds 0.3. Therefore, we use a default ratio of 0.3 in our main results to objectively assess the impact of harmful noise.

\subsection{The effects of Beneficial Noise under Different Noise Ratios}
To demonstrate the positive effects at different harmful noise ratios, we present comprehensive results for illegal sentence noise disturbances with ratios ranging from 0 to 0.8. As shown in Table \ref{table9}, we see that the introduction of illegal sentence noise (beneficial noise) consistently enhances model performance, thereby illustrating the generalization of beneficial noise.

\begin{table*}[!ht]
\centering
\caption{Additional results for different illegal sentence noise (ISN) ratios on RGB. L2-13B, L3-8B, Q2-7B, M-7B, V-13B, B2-13B, L3-70B and M-8x7B represents Llama2-13B, Llama3-8B-Instruct, Qwen2-Instruct, Mistral-7B-Instruct-v0.2, Vicuna-13B-v1.5, Baichuan2-13B-chat, Llama3-70B-Instruct, Mixtral-8x7B-Instruct.}
\label{table9}
\resizebox{1\textwidth}{!}{
\begin{tabular}{l|cccccc|cc|l}
\toprule
\multirow{2.5}{*}{{\bfseries Scenario}} & \multicolumn{6}{c|}{{\bfseries Small}} & \multicolumn{2}{c|}{{\bfseries Large}} & \multirow{2}{*}{{\bfseries Average}} \\
\cmidrule(lr){2-7} \cmidrule(lr){8-9}
\quad           & L2-13B & L3-8B & Q2-7B & M-7B & V-13B & B2-13B & L3-70B & M-8x7B & \\
\midrule
0               & 29.33  & 80.00 & 76.33 & 80.33 & 80.33 & 78.00 & 76.00 & 77.33 & 72.21 \\
\quad + ISN     & 72.33  & 83.00 & 80.00 & 81.00 & 82.33 & 79.67 & 79.67 & 79.67 & {\bfseries 79.71}$_{\textcolor{red}{+7.50}}$ \\
\midrule
0.2             & 18.67  & 77.33 & 75.33 & 76.00 & 79.33 & 73.33 & 76.67 & 73.67 & 68.79 \\
\quad + ISN     & 73.67  & 82.67 & 80.33 & 76.67 & 80.00 & 72.33 & 80.33 & 73.67 & {\bfseries 77.46}$_{\textcolor{red}{+8.67}}$ \\
\midrule
0.4             & 12.33  & 73.67 & 71.33 & 69.00 & 72.67 & 68.00 & 76.33 & 65.67 & 63.63 \\
\quad + ISN     & 70.67  & 77.00 & 73.00 & 71.00 & 73.33 & 68.33 & 80.00 & 66.67 & {\bfseries 72.50}$_{\textcolor{red}{+8.87}}$ \\
\midrule
0.6             & 8.67   & 72.33 & 66.00 & 65.33 & 67.00 & 63.67 & 82.00 & 64.33 & 61.17 \\
\quad + ISN     & 69.33  & 72.00 & 66.67 & 64.67 & 70.00 & 66.33 & 79.33 & 63.67 & {\bfseries 69.00}$_{\textcolor{red}{+7.33}}$ \\
\midrule
0.8             & 8.00   & 70.33 & 62.67 & 61.33 & 68.67 & 63.67 & 78.00 & 62.33 & 59.38 \\
\quad + ISN     & 68.33  & 70.67 & 64.67 & 63.33 & 69.00 & 66.33 & 78.33 & 63.67 & {\bfseries 68.04}$_{\textcolor{red}{+8.66}}$ \\
\bottomrule
\end{tabular}}
\end{table*}

\begin{table}[htp]
\centering
\caption{Additional control experiments by varying the repetition of answer-containing text chunks.\label{table11}}
\resizebox{1.0\linewidth}{!}{
\begin{tabular}{l|ccccl}
\toprule
& {\bfseries 512} & {\bfseries 748} & {\bfseries 1024} & {\bfseries 2048} & {\bfseries Avg}\\
\midrule
Golden only    & 79.80 & 83.00 & 83.20 & 83.40 & 82.35 \\
Golden \& CN   & 21.20 & 23.60 & 23.80 & 23.80 & 23.10 \\
Golden \& SuN  & 81.00 & 82.00 & 82.60 & 82.40 & 82.00 \\
Golden \& ON   & 78.60 & 79.80 & 80.00 & 80.00 & 79.60 \\
Golden \& SeN  & 87.20 & 89.20 & 89.00 & 89.20 & 88.65$_{\textcolor{red}{+6.30}}$ \\
Golden \& DN   & 82.00 & 85.60 & 85.80 & 86.20 & 84.90$_{\textcolor{red}{+2.55}}$ \\
Golden \& ISN  & 87.40 & 89.80 & 90.00 & 89.80 & 89.25$_{\textcolor{red}{+6.90}}$ \\
\bottomrule
\end{tabular}}
\end{table}

\subsection{Additional Control Experiments}
 We conducted additional control experiments by varying the repetition of answer-containing text chunks. We present the results on 2WikiMQA using Llama3-8B in Table~\ref{table11}. While these factors did have some impact on performance, our core finding—that RAG noise can be categorized into beneficial and harmful types—remains consistent.

\subsection{The Impact of RAG Noise on Other Tasks like Mathematical Reasoning}
Given that previous discussions focused on QA tasks, it remains unclear whether the beneficial noise affects other tasks. To address this, we conduct experiments on mathematical reasoning, which requires higher cognitive and reasoning abilities~\citep{guo2025deepseek,team2025kimi}. Following prior research, we apply the PAL methodology to evaluate reasoning results. This approach involves using LLMs to parse natural language problems, generate intermediary programmatic solutions, and execute these solutions via a Python interpreter.

\begin{table}[!ht]
\centering
\caption{Evaluation results (accuracy (\%) for mathematical reasoning using GPT-3.5-turbo as the base model. The four conditions are zero-shot without noise, two-shot without noise, and perturbations to numeric and operator elements in 2-shot examples. $\bigtriangleup$ denotes the accuracy improvement (\%) with noise compared to no noise.}\label{table12}
\resizebox{0.47\textwidth}{!}{
\begin{tabular}{l|ccc}
\toprule
{\bfseries Scenario} & {\bfseries GSM8K}& {\bfseries GSMHard}& {\bfseries Average ($\bigtriangleup$)}  \\
\midrule
0-shot            & 50.40 & 40.20 & 45.30 \\
2-shot-no-noise   & 55.40 & 47.80 & 51.60 \\
2-shot-num        & 65.40 & 50.60 & 58.00 (+6.40) \\
2-shot-operator   & 62.20 & 53.20 & 57.70 (+6.10) \\
\bottomrule
\end{tabular}}
\end{table}

As shown in Table \ref{table12}, introducing numeric or operator perturbations to retrieved examples significantly improves model performance (by 6.40$\%$ and 6.10$\%$, respectively). We hypothesize that this mechanism resembles adversarial training~\citep{wang2019improving}. Specifically, these perturbations likely help the model implicitly learn to identify and address potential errors or ambiguities, thereby enhancing its robustness. As a result, LLMs are better equipped to reason accurately amidst unclear or noisy test examples due to this implicit training. We anticipate that the insights presented in this paper could benefit other fields like creative writing, visual reasoning, and 3D generation~\citep{zhao2024retrievalaugmentedgenerationaigeneratedcontent,YIN2025104275,wu2025boosting,team2025kimi}.

\subsection{Detailed Statistical Validation}
To statistically evaluate the differences between scenarios with and without beneficial noise,, we apply the nonparametric Wilcoxon signed-rank test~\citep{wilcoxon1992individual}. This statistical test is specifically designed to compare two related samples or repeated measurements when the data may not follow a normal distribution, making it particularly suitable for our analysis.
The Wilcoxon signed-rank test evaluates whether there is a significant difference between paired observations through the following procedure:

\begin{enumerate}
\item \textbf{Calculate differences:} For each pair of values $X_i$ and $Y_i$, compute the difference $D_i = X_i - Y_i$.

\item \textbf{Rank differences:} Take the absolute values $|D_i|$ and rank them from smallest to largest, denoted as $R_i$. For ties, average ranks are assigned.

\item \textbf{Assign signs to ranks:} For each pair $(X_i, Y_i)$, assign the sign of $D_i$ to its corresponding rank: $R'_i = \text{sign}(D_i) \cdot R_i$, where $\text{sign}(D_i) = +1$ if $D_i > 0$, $-1$ if $D_i < 0$, and $0$ if $D_i = 0$.

\item \textbf{Calculate rank sums:} Separate the ranks into positive and negative sums: $W^+ = \sum_{D_i > 0} R'_i$ and $W^- = \sum_{D_i < 0} R'_i$.

\item \textbf{Determine test statistic:} The test statistic $W$ is the smaller of the two sums: $W = \min(W^+, W^-)$.

\item \textbf{Calculate p-value:} The p-value is derived from the distribution of the test statistic $W$.
\end{enumerate}

If the p-value is smaller than the chosen significance level (e.g., 0.05), the null hypothesis (that there is no difference between the paired samples) is rejected, indicating a statistically significant difference.

In our analysis, we test the null hypothesis of no significant difference ($H_0: \text{difference} = 0$) against the alternative hypothesis of a significant difference ($H_1: \text{difference} \neq 0$). Following common practice, we use a significance level of 0.05 (5e-2). Specifically, we use the Wilcoxon Signed-Rank Test to evaluate performance differences before and after introducing beneficial noise (e.g., ISN). Results in Table \ref{table4} in the main text confirm statistically significant improvements in model performance, highlighting the positive impact of beneficial noise.

\begin{table}[ht]
\caption{Attention distribution across documents in different scenarios}
\label{tab:attention_analysis}
\centering
\resizebox{0.47\textwidth}{!}{
\begin{tabular}{lccccc}
\toprule
\textbf{Scenario} & \textbf{Doc1} & \textbf{Doc2} & \textbf{Doc3} & \textbf{Doc4} & \textbf{Doc5} \\
\midrule
Golden \& CN & 0.54 & 0.34 & 0.28 & 0.49 & \textbf{0.76} \\
Golden Only & - & - & \textbf{1.00} & - & - \\
Golden \& ISN & 0.45 & 0.23 & \textbf{0.67} & 0.58 & 0.60 \\
\bottomrule
\end{tabular}}
\end{table}

\subsection{In-depth exploration of the underlying mechanisms}
To better understand the mechanisms behind RAG noise effects, we conduct an in-depth analysis of model attention patterns. Following previous studies~\citep{zhu2024promptrobustevaluatingrobustnesslarge,wu2024can}, we employ \textit{Attention by Gradient} as our visualization technique to examine how different noise types influence attention distribution across retrieved documents. Our analysis follows a three-step gradient-based approach:

\begin{enumerate}
    \item \textbf{Token-Level Gradient Computation:}\\
    For each token $t_{i,j}$, we calculate the gradient of the loss function $L$ (cross-entropy loss by default) with respect to the token:
    \begin{equation}
        g_{i,j} = \frac{\partial L(f_M(x), y)}{\partial t_{i,j}}
    \end{equation}
    where $f_M$ represents the model function, $x$ denotes the input, and $y$ is the target output.

    \item \textbf{Word-Level Gradient Aggregation:}\\
    We aggregate token-level gradients to obtain word-level attention scores by summing gradients corresponding to each word $w_i$:
    \begin{equation}
        g_w = \sum_{j=0,1,...,n} g_{i,j}, \quad \text{s.t.} \quad w_i = f_{\text{map}}(t_{i,j})
    \end{equation}
    
    \item \textbf{Document-Level Score Normalization:}\\
    Given our Top-5 retrieval setting, we aggregate word-level gradients into document-level attention scores and normalize them to the range $[0, 1]$ to facilitate cross-document comparison.
\end{enumerate}

We conduct a comparative analysis examining three scenarios: golden context only, golden context with beneficial noise (Illegal Sentence Noise, ISN), and golden context with harmful noise (Counterfactual Noise, CN). Table~\ref{tab:attention_analysis} presents the attention distribution results, where \texttt{Doc3} represents the golden context and the remaining documents (\texttt{Doc1}, \texttt{Doc2}, \texttt{Doc4}, \texttt{Doc5}) contain various noise types. The document receiving the highest attention in each scenario is highlighted in bold.

We have two key findings:

\begin{itemize}
    \item \textbf{Enhanced Focus on Essential Information:} Beneficial noise, such as Illegal Sentence Noise (ISN), helps the model focus on critical content. In Table~\ref{tab:attention_analysis}, the golden context (\texttt{Doc3}) retains the highest attention score under Golden \& ISN, suggesting that beneficial noise guides the model to prioritize relevant details, improving reasoning and accuracy.
    
    \item \textbf{Diminished Impact of Misleading Noise:} Harmful noise, like Counterfactual Noise (CN), misdirects the model's attention to irrelevant or incorrect documents. This diversion reduces the model's ability to focus on the golden context, leading to a performance drop.
\end{itemize}

These findings provide empirical evidence for the dual nature of noise in RAG systems. We leave more attention visualization across different model layers and extensive gradient flow analysis for future investigation.

\subsection{Case Study}
Table \ref{table13}-\ref{table16} presents the reasoning process of Llama3-8B-instruct. Without beneficial noise, the model ignores correct information and exhibits logical flaws under counterfactual noise influence. However, upon introducing beneficial noise, the model exhibits heightened attention to the golden context and successfully distinguishes between correct and incorrect information.

\begin{table*}[ht]
\centering
\caption{Example $\#$1 of LLM outputs on the single-hop dataset without and with beneficial noise (BN). The {\bfseries \textcolor{blue}{blue}} and {\bfseries \textcolor{green}{green}} colors denote the correct and incorrect responses, respectively. `GT' denotes ground truth. Upon introducing BN, LLMs exhibit clearer reasoning processes, more standardized response formats, and increased focus on golden context.}\label{table13}
\begin{tabular}{p{1.5cm}|p{13.6cm}}
\toprule
{\bfseries Retrieved Context \quad wo BN} & 
[1] Paul R. Ehrlich, author of "The Population Bomb," was born on \textcolor{red}{June 14, 1931}. Ehrlich, a renowned biologist, published his influential book on population control in 1968, which sparked $\ldots$ {\bfseries\textcolor{red}{[Counterfactual]}}\newline
[2] The author of The Population Bomb, Paul Ehrlich, was born on \textcolor{red}{May 29, 1932}. He is an American biologist and educator, known for his warnings about the consequences of overpopulation $\ldots$  \,\quad\quad{\bfseries\textcolor{red}{ [Golden Context]}}\\
\midrule
{\bfseries Question} & 
When was the author of The Population Bomb born?\newline
`A': June 14, 1931\quad`B': Uncertain\quad\textcolor{red}{`C': May 29, 1932 (Ground Truth)}\quad`D': April 22, 1934 \\

\midrule
{\bfseries Answer\quad wo BN} & {\bfseries \textcolor{green}{A}}  $\#\#\#$\textbackslash n\textbackslash n$\#\#\#$ Explanation:\textbackslash n The correct answer is `A': June 14, 1931. According to the information provided, Paul R. Ehrlich, the author of ``The Population Bomb'' was born on \textcolor{green}{June 14, 1931}. The other options are incorrect, as they provide different birth dates for the author. \quad \textcolor{red}{\faTimes}\\
\midrule
{\bfseries Answer\quad w BN} & {\bfseries \textcolor{blue}{C}}\textbackslash n$\#\#\#$ Explanation: According to information [2], Paul Ehrlich, the author of The Population Bomb, was born on \textcolor{blue}{May 29, 1932}. Therefore, the correct answer is `C'. Information [1] provides a different birthdate, but it is not the correct answer. The other options are not mentioned in the given information.\quad \textcolor{red}{\faCheck}\\
\bottomrule
\end{tabular}
\end{table*}

\begin{table*}[htp]
\centering
\caption{Example $\#$2 of LLM outputs on the single-hop dataset without and with beneficial noise (BN). The {\bfseries \textcolor{blue}{blue}} and {\bfseries \textcolor{green}{green}} colors denote the correct and incorrect responses, respectively. `GT' denotes ground truth. Upon introducing BN, LLMs exhibit clearer reasoning processes, more standardized response formats, and increased focus on golden context.}\label{table14}
\begin{tabular}{p{1.5cm}|p{13.6cm}}
\toprule
{\bfseries Retrieved Context \quad wo BN} & 
[1] \textcolor{red}{Northrop Corporation} built the fastest air-breathing manned aircraft, the Northrop Grumman B-2 Spirit. Known for its stealth technology and high-speed capabilities, it represents a pinnacle in modern aviation technology. {\bfseries\textcolor{red}{[Counterfactual]}}\newline
[2] The \textcolor{red}{Lockheed Corporation} built the fastest air-breathing manned aircraft, the Lockheed SR-71 Blackbird. This reconnaissance aircraft holds the record for the fastest speed ever recorded by a manned aircraft. \,\quad\quad{\bfseries\textcolor{red}{ [Golden Context]}}\newline
$\ldots$\\
\midrule
{\bfseries Question} & 
Who built the fastest air-breathing manned aircraft?\newline
`A': Uncertain\quad`B': Boeing Corporation\quad\textcolor{red}{`C': Lockheed Corporation (GT)}\quad`D': Northrop Corporation \\
\midrule
{\bfseries Answer\quad wo BN} & {\bfseries \textcolor{green}{D}}\textbackslash n \textbackslash n$\#\#\#$ Explanation:\textbackslash n The information provided states that Northrop Corporation built the Northrop Grumman B-2 Spirit, which is known for its stealth technology and high-speed capabilities. However, it does not explicitly state that this aircraft is the fastest air-breathing manned aircraft. \quad \textcolor{red}{\faTimes} \\
\midrule
{\bfseries Answer\quad w BN} & {\bfseries \textcolor{blue}{C}}\textbackslash n$\#\#\#$ Explanation: The information provided states that the Lockheed Corporation built the Lockheed SR-71 Blackbird, which is the fastest air-breathing manned aircraft. Therefore, the answer is 'C'.\quad \textcolor{red}{\faCheck}\\
\bottomrule
\end{tabular}
\end{table*}

\begin{table*}[htp]
\centering
\caption{Example $\#$3 of LLM outputs on the multi-hop dataset without and with beneficial noise (BN). The {\bfseries \textcolor{blue}{blue}} and {\bfseries \textcolor{green}{green}} colors denote the correct and incorrect responses, respectively. Upon introducing BN, LLMs exhibit clearer reasoning processes, more standardized response formats, and increased focus on golden context.}\label{table15}
\begin{tabular}{p{1.5cm}|p{13.6cm}}
\toprule
{\bfseries Retrieved Context \quad wo BN} & 
[1] \textcolor{red}{Robert Harley}, Earl of Oxford and Earl Mortimer, was the paternal grandfather of Edward Harley, 5th Earl of Oxford and Earl Mortimer. He was a prominent statesman and served as the Lord High Treasurer of Great Britain. {\bfseries\textcolor{red}{[Counterfactual]}}\newline
[2] John Harley (29 September 1728 – 7 January 1788) was a British bishop. Harley was the second son of \textcolor{red}{Edward Harley}, 3rd Earl of Oxford and Earl Mortimer. He was Archdeacon of Shropshire from 1760 to 1769 and then Archdeacon of Hereford from 1769 to 1787. {\bfseries\textcolor{red}{ [Golden Context]}}\newline
$\ldots$\\
\midrule
{\bfseries Question} & 
Who is the paternal grandfather of Edward Harley, 5Th Earl Of Oxford And Earl Mortimer?\newline
\textcolor{red}{`A': Edward Harley, 3rd Earl of Oxford and Earl Mortimer (GT)}\quad`B': Uncertain\quad`C': Uncertain\quad`D': Thomas Harley, 4th Earl of Oxford and Earl Mortimer \\
\midrule
{\bfseries Answer\quad wo BN} & {\bfseries \textcolor{green}{C}}  \textbackslash n\textbackslash n The paternal grandfather of Edward Harley, 5th Earl of Oxford and Earl Mortimer, is Robert Harley, Earl of Oxford and Earl Mortimer. Therefore, the correct answer is 'C'. \quad \textcolor{red}{\faTimes}\\
\midrule
{\bfseries Answer\quad w BN} & {\bfseries \textcolor{blue}{A}}\textbackslash n\textbackslash n$\#\#\#$ Explanation: nThe information provided states that John Harley (1728-1788) was the second son of Edward Harley, 3rd Earl of Oxford and Earl Mortimer. Therefore, Edward Harley, 3rd Earl of Oxford and Earl Mortimer is the paternal grandfather of Edward Harley, 5th Earl of Oxford and Earl Mortimer.\quad \textcolor{red}{\faCheck}\\
\bottomrule
\end{tabular}
\end{table*}

\begin{table*}[htp]
\centering
\caption{Example $\#$4 of LLM outputs on the multi-hop dataset without and with beneficial noise (BN). The {\bfseries \textcolor{blue}{blue}} and {\bfseries \textcolor{green}{green}} colors denote the correct and incorrect responses, respectively. Upon introducing BN, LLMs exhibit clearer reasoning processes, more standardized response formats, and increased focus on golden context.}\label{table16}
\begin{tabular}{p{1.5cm}|p{13.6cm}}
\toprule
{\bfseries Retrieved Context \quad wo BN} & 
[1] With an estimated base length of 230 meters, 200 men end to end would only cover about 46$\%$ of the Great Pyramid of Giza's base, highlighting the massive scale and size of the ancient structure. {\bfseries\textcolor{red}{[Counterfactual]}}\newline
[2] The base of the Great Pyramid of Giza is 756 feet long. The average height of a man is 5 foot 9.  \,\quad\quad{\bfseries\textcolor{red}{ [Golden Context]}}\newline
$\ldots$\\
\midrule
{\bfseries Question} & 
Can 200 men end to end cover Great Pyramid of Giza's base?\newline
`A': Uncertain\quad\textcolor{red}{`B': False (GT)}\quad`C': True \\
\midrule
{\bfseries Answer\quad wo BN} & {\bfseries \textcolor{green}{C}} \quad \textcolor{red}{\faTimes}\\
\midrule
{\bfseries Answer\quad w BN} & {\bfseries \textcolor{blue}{B}}\textbackslash n\textbackslash n$\#\#\#$ Information:\textbackslash n The base of the Great Pyramid of Giza is 756 feet long. The average height of a man is 5 foot 9. (int) The number of men required to cover the base end to end is 756 / 5.75.\quad \textcolor{red}{\faCheck}\\
\bottomrule
\end{tabular}
\end{table*}

\end{document}